%% file: main.tex
\documentclass[lettersize,journal]{IEEEtran}
\usepackage{amsmath,amsfonts}
\usepackage{algorithmic}
\usepackage{algorithm}
\usepackage{array}
\usepackage[caption=false,font=normalsize,labelfont=sf,textfont=sf]{subfig}
\usepackage{textcomp}
\usepackage{stfloats}
\usepackage{url}
\usepackage{verbatim}
\usepackage{graphicx}
\usepackage{cite}

\begin{document}

\title{Separability Membrane: 3D Active Contour for Point Cloud Surface Reconstruction}
\author{Gulpi Qorik Oktagalu Pratamasunu, Guoqing Hao,
        Kazuhiro Fukui\IEEEmembership{, Member, IEEE}%
\thanks{Manuscript received February XX, 2025; revised May XX, 2025.}
\thanks{Gulpi Qorik Oktagalu Pratamasunu and Kazuhiro Fukui are with Department of Computer Science, Graduate School of Systems and Information Engineering, University of Tsukuba, Tsukuba 305-0085, Japan.}
\thanks{Guoqing Hao is with Department of Integrated Information Technology, Aoyama Gakuin University, Sagamihara Kanagawa, 252-5258, Japan.}
}

\markboth{IEEE -,~Vol.~XX, No.~X, February~2025}%
{Pratamasunu \MakeLowercase{\textit{et al.}}: Separability Membrane: 3D Active Contour for Point Cloud Surface Reconstruction}

\IEEEpubid{0000--0000/00\$00.00~\copyright~2021 IEEE}

\maketitle

\begin{abstract}

This paper proposes Separability Membrane, a robust 3D active contour for extracting a surface from 3D point cloud object. Our approach defines the surface of a 3D object as the boundary that maximizes the separability of point features, such as intensity, color, or local density, between its inner and outer regions based on Fisher's ratio. Separability Membrane identifies the exact surface of a 3D object by maximizing class separability while controlling the rigidity of the 3D surface model with an adaptive B-spline surface that adjusts its properties based on the local and global separability. A key advantage of our method is its ability to accurately reconstruct surface boundaries even when they are ambiguous due to noise or outliers, without requiring any training data or conversion to volumetric representation. Evaluations on a synthetic 3D point cloud dataset and the 3DNet dataset demonstrate the membrane's effectiveness and robustness under diverse conditions.
\end{abstract}


\input{Chapters/1_intro_point_cloud}
\input{Chapters/2_new}
\input{Chapters/3_proposed}

\input{Chapters/4_experiments}
\input{Chapters/5_conclusion}


\bibliography{main.bib}{}
\bibliographystyle{IEEEtran}
 
\vspace{11pt}

\begin{IEEEbiography}[{\includegraphics[width=1in,height=1.25in,clip,keepaspectratio]{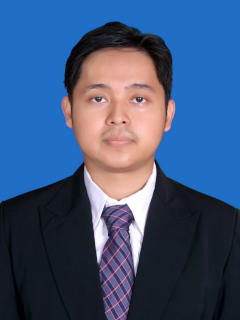}}]{Gulpi Qorik Oktagalu Pratamasunu} received the M.S. degree in Computer Science from the Department of Informatics, Institut Teknologi Sepuluh Nopember, Indonesia, in 2015. He is a lecturer with Department of Informatics, Universitas Nurul Jadid, Indonesia and currently pursuing the Ph.D. degree with the Department of Computer Science, University of Tsukuba, Japan. His research interests include image processing, computer vision, and computer graphics.
\end{IEEEbiography}

\begin{IEEEbiography}[{\includegraphics[width=1in,height=1.25in,clip,keepaspectratio]{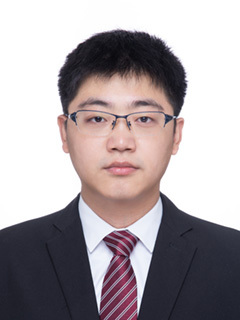}}]{Guoqing Hao} received the Ph.D. degree in Engineering from University of Tsukuba, Japan, in 2024. He is currently an Assistant Professor with the Department of Integrated Information Technology, Aoyama Gakuin University. His research interests include image editing, computer vision, and machine learning.
\end{IEEEbiography}

\begin{IEEEbiography}[{\includegraphics[width=1in,height=1.25in,clip,keepaspectratio]{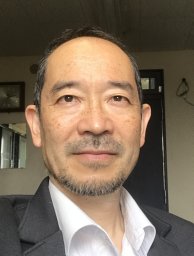}}]{Kazuhiro Fukui} (Member, IEEE) received the Ph.D. degree from the Tokyo Institute of
Technology, in 2003. He joined the Toshiba Corporate Research and Development Center. He was
a Senior Research Scientist with the Multimedia
Laboratory. He is currently a Professor with
the Department of Computer Science, Graduate
School of Systems and Information Engineering,
University of Tsukuba. His research interests
include the theory of machine learning, computer
vision, pattern recognition, and their applications. He is a member of the
SIAM.
\end{IEEEbiography}

\end{document}

%% file: Chapters/1_intro_point_cloud.tex
\section{Introduction}

\IEEEPARstart{S}{urface} reconstruction from point clouds is a vital process in various fields, including computer vision and computer graphics. Point clouds, which consist of collections of 3D points in space, provide a fundamental way to represent 3D objects. The primary objective of surface reconstruction is to generate a continuous and smooth surface representation from these discrete sets of points.
This reconstructed surface serves as a crucial foundation for numerous real-world applications, ranging from digital cultural heritage \cite{Alkadri2022Investigating, Moyano2021Operability, Fryskowska2018A} to quality control and inspection in the industry \cite{defect_detection_multi, Liu2023Near, Wang2019Computational, Wang2019Applications}. Furthermore, incorporating reconstructed surface data as additional input has been shown to improve the accuracy of downstream tasks such as action recognition, classification, and part segmentation by providing a more structured and continuous representation of objects \cite{action_recognition, quadratic}.

There are several critical challenges in the process of reconstructing surfaces from point cloud data. Real-world point cloud data often lacks completeness due to occlusions during scanning. Sensor inaccuracies and misalignment introduce noise in point positions. Measurement errors create significant outliers that affect reconstruction accuracy. Additionally, non-uniform sampling densities across surfaces make it difficult to preserve fine geometric details.

Traditional approaches for surface reconstruction are computationally efficient, but they have limitations in handling these challenges. For instance, Alpha-Shapes \cite{Edelsbrunner1983} uses geometric parameters to create simplified shapes; however, they are sensitive to the choice of parameters and struggle with non-uniform sampling. Poisson Surface Reconstruction (PSR) \cite{Kazhdan2006} is robust against noise through implicit function fitting but depends heavily on accurate normal estimation, making it vulnerable to outliers and noisy data. Additionally, the conventional methods mainly focus on geometric information while often neglecting other salient features such as intensity values, color attributes, and local density distributions.

Recently, deep learning-based methods have emerged to address these limitations. Approaches such as Point2Mesh \cite{point2mesh}, ParSeNet \cite{parsenet}, Shape-as-Points \cite{Peng2021SAP}, and BPNet framework \cite{bpnet, bezier} use learned priors for improved reconstruction quality and noise resistance. However, these methods need a significant amount of training data and can be computational complex. Furthermore, they often fail to generalize to novel geometric shapes outside their training distribution. These challenges highlight the need to explore alternative approaches that can perform reconstruction without extensive training data while maintaining robustness to noise, outlier, and sampling irregularities.

\begin{figure}[t]
    \centering
    \includegraphics[width=8.5cm]{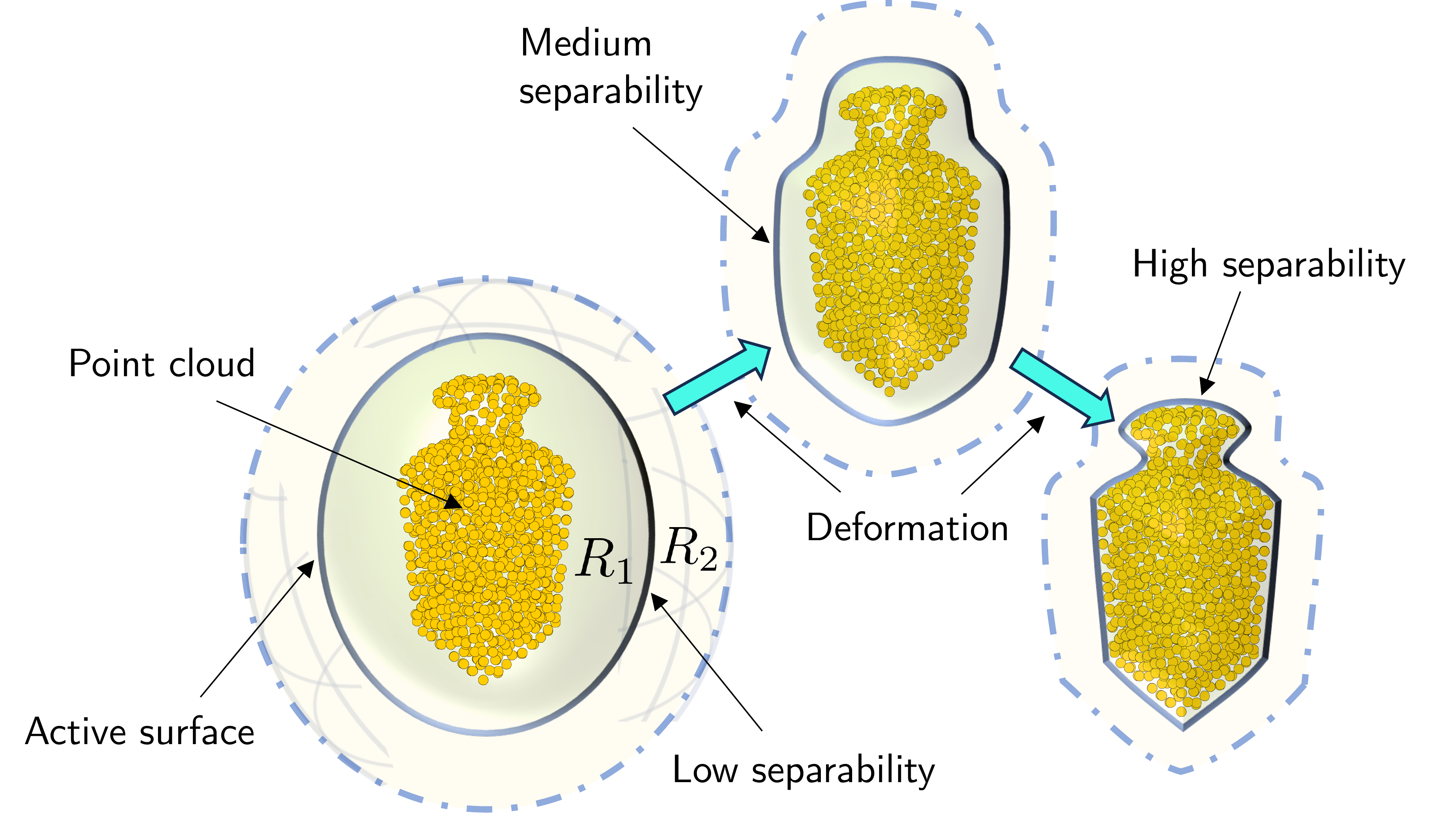}
    \caption{Basic idea of finding the surface of a 3D point cloud object based on class separability. Our method identifies the object boundary by gradually deforming the active surface while maximizing the class separability between the inner volume region ($R_1$) and the outer volume region ($R_2$).}
    \label{fig:sep_mem_overview}
\end{figure}

Active contour models, including traditional snakes \cite{kass1988snakes} and level sets \cite{caselles1995geodesic, chan2001active}, offer a promising alternative framework. They were originally developed for extracting a 2D contour of an object in an image, leading to a task of 2D image segmentation. Despite their success in 2D and 3D image \IEEEpubidadjcol segmentation, extending active contours to point cloud data presents unique challenges.

Traditional active contour models generally rely on volumetric data representations, such as image intensity from an RGB camera, CT images, and ultrasonic images. Consequently, they struggle to effectively handle the irregular nature of point cloud data. Previous research has attempted to tackle this issue by converting irregular point cloud data to a grid-like representation\cite{voxnet2015, Yataka2017FeaturePE}. However, this process often results in unnecessary data redundancy and quantization artifacts\cite{pointnet}. 

In this paper, we propose the {\it Separability Membrane}, an active surface model which directly utilizes raw point cloud data for surface reconstruction.  Here, we define the object's surface as the boundary that maximizes the class separability between inner and outer regions ($R_1$ and $R_2$), as shown in Fig. \ref{fig:sep_mem_overview}. Given the initial surface, our membrane gradually deforms its shape while maximizing the separability between two volume regions, inside and outside of the target 3D object, without converting points to volumetric representation.

To achieve this, we propose a method for calculating the class separability directly from a given point cloud data. Instead of transforming the data into a binary occupancy grid \cite{voxelization, voxnet2015}, we assign binary values directly in the point cloud space, 1 for data points and 0 for non-data points (uniformly distributed virtual points that represent the unoccupied space). This naive method works as expected when the data distribution is comparatively uniform. However, it does not work effectively for an inhomogeneous data distribution, as it generate unbalanced dataset and inaccurate degree of separability calculation.

To address this problem, we introduce a mechanism for estimating the distribution of non-data points based on the local density of the given data points. The basic idea behind this function is that the distribution of non-data points should depend on the local density of given data. In other words, regions of high data density require more non-data points, while sparse regions need fewer. This function enables us to generate an optimal set of non-data points, resulting in stable and accurate separability calculations for any data distribution. Note that although we discussed the distributions of data and non-data, we need to consider only the number of non-data since we can calculate the degree of separability without the spatial information of each data.

We represent the membrane using a cubic B-spline surface, inspired by the B-spline-based active contour model\cite{cipolla1990dynamic, wakasugi2004robust, cipolla1992surface} where a B-spline curve represents a contour model.
The stiffness of the surface model depends on the number of control points used to represent the whole surface. A sparse control point distribution creates a rigid surface, while a dense distribution makes the membrane more flexible, allowing the surface model to capture fine geometric details. However, controlling a highly flexible surface model is a difficult task.
To address this problem, we introduce a dynamic membrane adjustment function to optimize the fitting process by dynamically changing the number of control points while maintaining reconstruction accuracy. 
This function enables the membrane to begin in a rigid state and gradually become more flexible as the process progresses, balancing computational efficiency with the ability to capture fine surface details. 

We evaluated the effectiveness of the proposed method through extensive experiments on synthetic and real-world point-cloud data. The results on a synthetic 3D dataset show the validity of our point separability in extracting meaningful features directly from raw point cloud data. Furthermore, the results on the 3DNet dataset \cite{3dnet} on various extreme conditions show the effectiveness of our method in the point-cloud surface reconstruction task.

Our main contributions are summarized as follows:
\begin{itemize}
\item[1)] A novel active surface model, {\it Separability Membrane}, that effectively extracts the surface of 3D objects from point cloud data in an unsupervised manner.
\item[2)] We introduce a new class separability, {\it Point Separability}, which works directly with raw point cloud data by considering spatial positions and other point attributes, enhancing the efficiency, flexibility and accuracy of the analysis.
\item[3)] A dynamic membrane function that adjusts the membrane's properties during iteration, balancing computational efficiency and detail representation.
\end{itemize}

The paper is organized as follows.
In Section II, we describe the related methods.
In Section III, we discuss the proposed method. First, we describe the extension of class separability concept to point cloud. Then, we explain the basic idea of our separability membrane. Further, we introduce a dynamic membrane properties adjustment.
In Section IV, we demonstrate the effectiveness of our method through evaluation experiments.
Finally, Section V concludes the paper.

%% file: Chapters/2_new.tex
\section{Related Work}
\label{sec:relatedwork}

In this section, we review three fundamental concepts that are central to our paper: point cloud surface reconstruction, active contour models, and class separability. 

\subsection{Point Cloud Surface Reconstruction}

Surface reconstruction has evolved significantly, encompassing various approaches to address the challenges of converting discrete 3D points into continuous surfaces. These challenges include handling noise, outliers, non-uniform sampling, and incomplete data while maintaining geometric fidelity and computational efficiency. The field has seen diverse methodological developments, from traditional geometric approaches to modern learning-based techniques.

Traditional methods can be broadly categorized into implicit surface methods and mesh-based approaches. Implicit surface methods, exemplified by Poisson Surface Reconstruction \cite{Kazhdan2006}, represent the surface as a level set of a continuous scalar field by solving a spatial Poisson equation. While effective at creating watertight surfaces, these methods require accurate normal estimates and struggle with large missing regions. Mesh-based approaches like Alpha shapes \cite{Edelsbrunner1983} and Ball-pivoting \cite{Bernardini1999} directly generate triangular meshes from point clouds, but their performance heavily depends on parameter selection and point density uniformity.

Recent learning-based methods have introduced new paradigms for surface reconstruction. Neural implicit representations \cite{mescheder2019occupancy} learn continuous signed distance functions or occupancy fields from point clouds, showing promising results in completing missing regions despite requiring significant training data. Point2Mesh \cite{point2mesh} combines learning-based features with traditional mesh deformation, though it can struggle with complex topologies. Shape as Points \cite{Peng2021SAP} bridges traditional geometric methods with modern learning techniques through a differentiable Poisson solver, yet faces challenges in preserving fine details and computational efficiency for large-scale reconstructions.

Learning-based primitive segmentation methods offer a different paradigm for surface reconstruction. Early methods focused on detecting specific primitive types, such as networks specialized in cuboid detection \cite{abstractionTulsiani17}. SPFN \cite{Li2019SPFN} expanded this capability by introducing supervised learning for detecting multiple canonical primitives including planes, spheres, cylinders, and cones, while ParSeNet \cite{parsenet} further added support for B-spline patches. BPNet \cite{bpnet, bezier} advances these approaches by introducing a more generic framework using Bezier patches, enabling unified processing of different primitive types. However, while BPNet's general formulation offers improved efficiency over methods that treat primitives separately, it can generate overly complex segmentations due to the nature of Bezier decomposition.

While these learning-based methods show promise, they share common limitations in requiring extensive training data and struggling with generalization to novel shapes. In contrast to these approaches, our work introduces a fundamentally different paradigm based on class separability between inner and outer regions of point clouds in an unsupervised manner. Unlike previous methods that rely on normal estimation estimation or chamfer distance computation, our membrane directly processes raw point cloud data without requiring training data or primitive assumptions. This approach enables us to robustly handle various point attributes while maintaining computational efficiency through dynamic membrane adjustment, addressing key challenges in point cloud surface reconstruction.

\subsection{Various Types of Active Contour Models}
The active contour model, also known as "snake", was introduced by Kass et al. \cite{kass1988snakes} as a valid approach to extract the contour of an object flexibly. The active contour model tries to search out the contour by minimizing an energy function that consists of internal energy (controlling smoothness and rigidity) and external energy derived from the image \cite{kass1988snakes}. In the energy minimization, the model modifies its shape gradually toward the contour.

The traditional snakes are effective for extracting the shape contours of an object. However, they face challenges when handling the topological problem on an object with multiple regions. The level set method was proposed to address this limitation by representing the contour implicitly as the zero level set of a higher-dimensional scalar function and evolving it based on partial differential equations \cite{caselles1995geodesic}. This implicit representation allows the contour to naturally adapt to changes in complicated topology, such as merging or splitting, making it a valid framework for modern active contour models. However, it is computationally expensive, sensitive to initialization, and requires careful parameter tuning when considering complex or noisy images as input.

Active contour models can be categorized based on their external energy formulation. In the edge-based active contour models \cite{kass1988snakes, caselles1995geodesic}, the external energy is derived from image gradients. These models works well for sharp contours, although it does not work stably with blurred contours. To address this, Chan and Vese \cite{chan2001active} developed the region-based active contours without calculating the edge strength from image intensity gradient. The region-based model is more robust and can handle images with weak or blurred contours. However, its reliance on the assumption of homogeneous regions makes it less practical for objects with heterogeneous or textured interiors. 

The separability snake \cite{wakasugi2004robust} offers an alternative approach to overcome the above issues. Instead of relying on image gradient or the assumption of homogeneous regions, this approach replaces them with a class separability degree that is calculated between its inner and outer regions. This technique provides more stable performance in weak boundary and noisy conditions by maximizing the class separability between the two regions using the Fisher ratio \cite{otsu1979threshold}. Furthermore, unlike previous methods, the active contour model is compactly represented using B-spline curves \cite{brigger2000b}, offering a more flexible and computationally manageable approach.

These advancements in the separability snake make them versatile and robust for image segmentation tasks. However, extending its framework to 3D active surfaces for 3D cloud point data can be challenging due to the limitation of the previous definition of separability degree to handle irregular data and the high computational cost of the fitting process in 3D space. 

\subsection{Image Analysis based on Class Separability}

\begin{figure}[t]
    \centering
    \includegraphics[width=7.0cm]{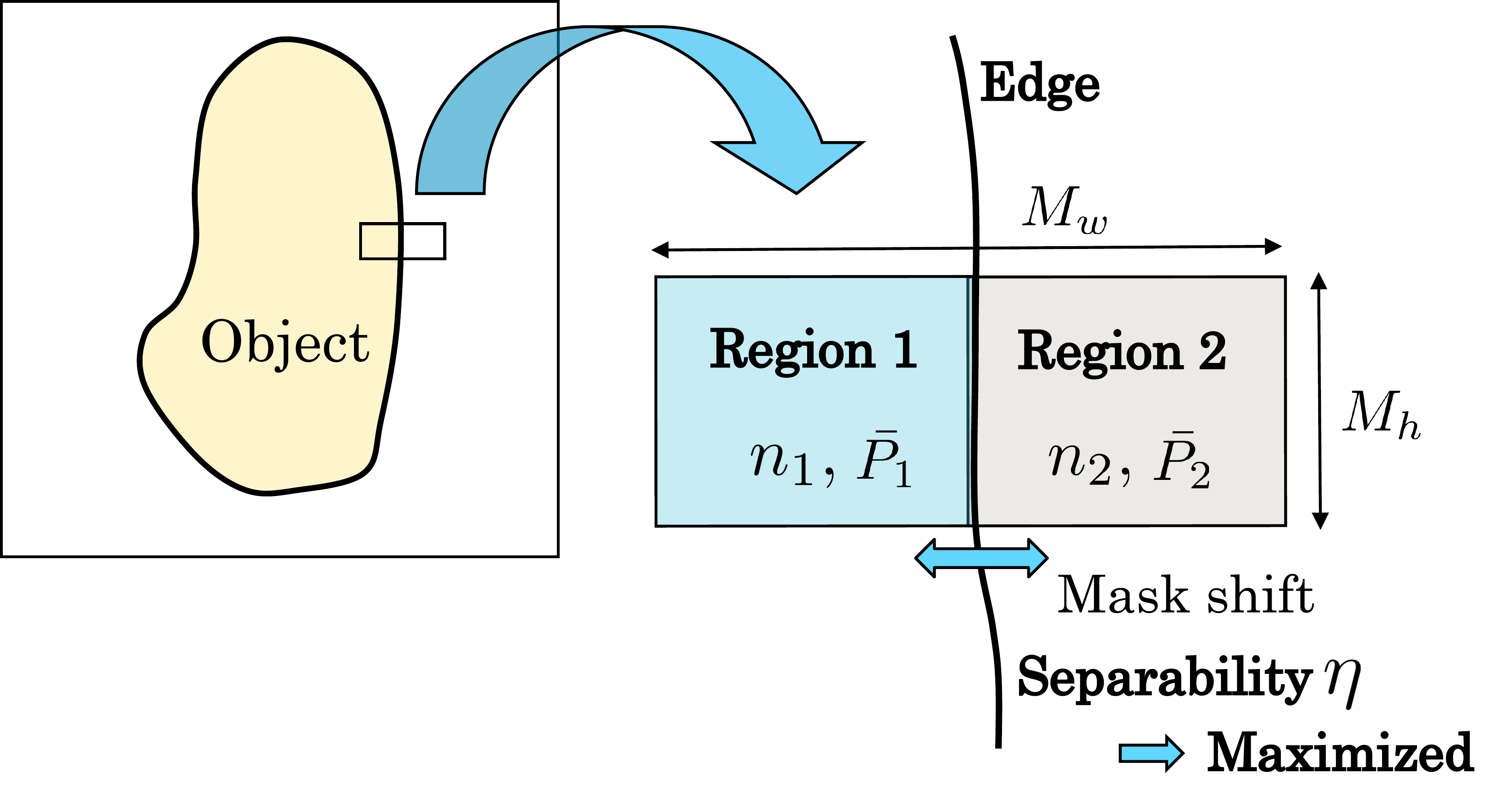}
    \caption{Definition of edge based on separability. Unlike traditional methods that rely on pixel intensity gradients, this approach defines edges by statistically analyzing the separability between adjacent regions.}
    \label{fig:sep_edge}
\end{figure}

There are various methods for image analysis based on the concept of class separability between the information from two regions. In image processing, the class separability degree was first introduced to a task of edge extraction \cite{fukui1995edge}. In this edge extraction, an edge was defined as not a boundary where a significant change in image intensity occurs but as a region boundary that separates two regions.
The edge filter based on separability degree can detect unclear and blurred edges more precisely and stably compared to edge filters based on image intensity gradient, as the separability degree can be stably calculated from the information of two local regions as shown in Fig. \ref{fig:sep_edge}.

Class separability between the regions, $\eta$, can be calculated from  two sets of image intensities $P_{f}$ from two local regions as follows
\begin{equation} \label{eq:separability}
\eta = \frac{\sigma_b^2}{\sigma_T^2}, 
\end{equation}
\begin{equation}
\sigma_b^2 = n_1(\overline{P_1}-\overline{P_m})^2 + n_2(\overline{P_2}-\overline{P_m})^2,
\end{equation}
\begin{equation}
\sigma_T^2 = \sum_{i=1}^N (P_i-\overline{P_m})^2,
\end{equation}
where $n_1$ and $n_2$ are the number of pixels in regions 1 and 2, $N$ is $n_1 + n_2$, $P_i$ is the image intensity in position $i$, $\overline{P_1}$ and $\overline{P_2}$ are the average image intensity values in regions 1 and 2, and $\overline{P_m}$ is the average image intensity of the entire region.

The concept of class separability has demonstrated significant utility across multiple computational domains. In finger shape recognition applications, this concept enables precise fingertip detection from RGB-D images \cite{Yataka2017FeaturePE}. In medical imaging applications, it enhances edge detection, particularly for intravascular ultrasound and tongue contour extraction \cite{liu2012ultrasound,liu13h_interspeech}. Furthermore, the methodology excels in detecting circular objects within noisy environments \cite{niigaki2012circular, ohkawa2011fast}, while specialized applications such as the Iris-Eyelid separability filter have advanced ocular tracking systems \cite{CHEN2013iriseyelid,chen2013irisoutline}. Applications in road feature extraction from monocular images \cite{kikuchi2014road} and pupil diameter measurement using visible-light cameras \cite{morita2016pupil} further demonstrate the approach's broad applicability.

%% file: Chapters/3_proposed.tex
\section{Proposed Method}
\label{sec:proposed}

Our method consists of three main components: (1) a novel separability approach that handles raw irregular point cloud data, (2) a weighted separability calculation for multiple attributes, and (3) an active surface model based on the local and global separability. These components work together to enable robust segmentation of 3D point cloud data without requiring conversion to regular grid structures.

\subsection{Non-data Points Estimation}

Three-dimensional point clouds inherently provide a direct representation of spatial structures with irregularly distributed points in space. While the concept of separability filter was originally developed for 2D image processing to detect edges based on the separability of image features rather than intensity gradients \cite{fukui1995edge}, extending this concept to 3D point clouds presents unique challenges. Prior works on 3D separability filter mainly rely on volumetric representations, requiring the depth information from RGB-D images to be converted into voxel grids \cite{Yataka2017FeaturePE}. Raw point cloud data transformed to volumetric representations using occupancy functions\cite{voxnet2015}, where voxels containing points are marked as 1 and empty voxels as 0. However, this transformation introduces unnecessary data redundancy and quantization artifacts \cite{pointnet}.

\begin{figure}[t]
    \centering
    \includegraphics[width=8.0cm]{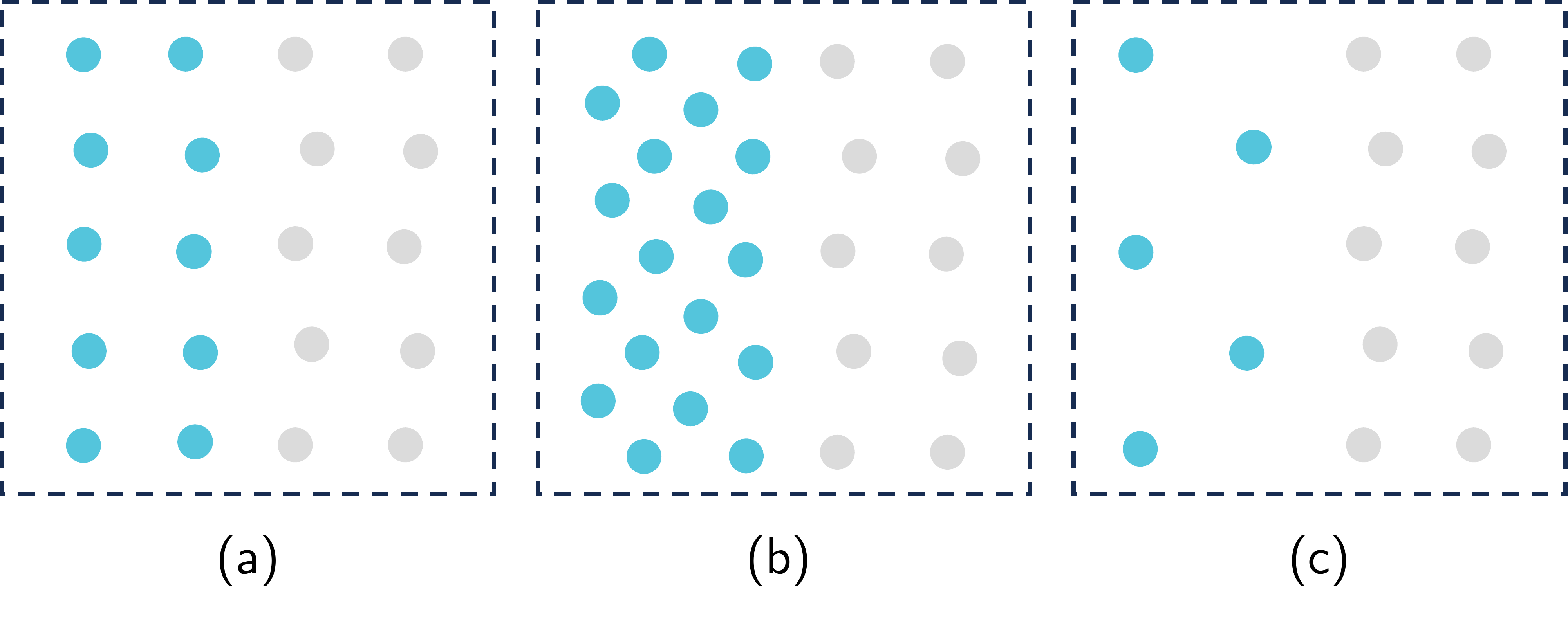}
    \caption{Non-data points estimation scenarios. (a) Ideal case: non-data points (gray points) match the local density of actual points (blue points). (b) Undersampling case: insufficient non-data points in dense regions. (c) Oversampling case: excessive non-data points in sparse regions.}
    \label{fig:density_compare}
\end{figure}

We address these problems by uniformly adding the non-data points to represent the empty space. In this way, with the exact number of non-data points, we can simulate the grid structure of volumetric representation directly in point cloud space, as shown in Fig. \ref{fig:density_compare}(a). However, choosing the appropriate number of non-data points of any given region is a challenging task. Too many non-data points will lead to oversampling, while insufficient non-data points will results in undersampling, as illustrated in Fig. \ref{fig:density_compare}(b) and (c).

\begin{figure}[t]
    \centering
    \includegraphics[width=8.0cm]{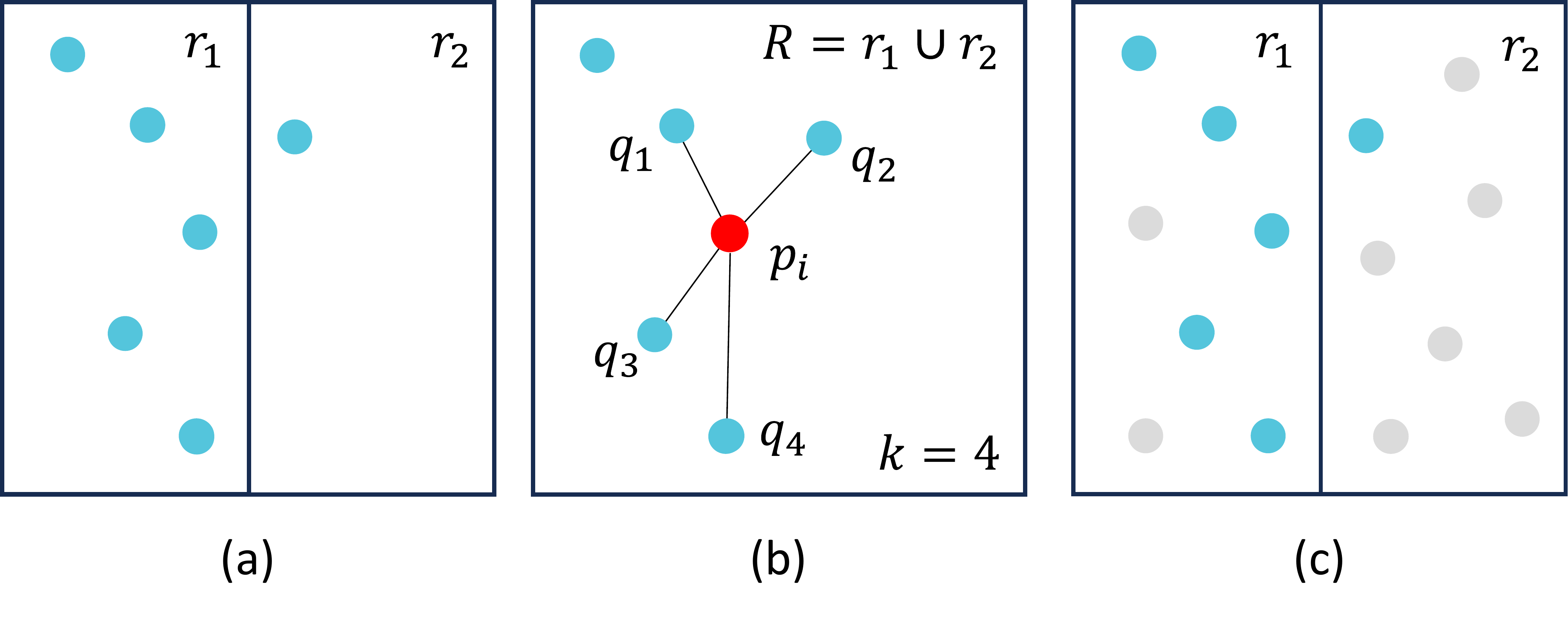}
    \caption{Adaptive non-data points estimation. (a) Adjacent regions containing actual points from point cloud data. (b) Local density estimation based on k-nearest neighbors on the merged region (c) Visualization of how non-data points would uniformly fill the empty space based on its local density. Note that non-data points are not explicitly added in the implementation, but their count is used in calculations.}
    \label{fig:point_types}
\end{figure}

To address this problem, we introduce an adaptive density estimation approach that accounts for local point distributions. When computing separability between two adjacent regions $r_1$ and $r_2$, we consider their union $R = r_1 \cup r_2$ for the local point density estimation, as shown in Fig. \ref{fig:point_types}. For each region $r \in {r_1, r_2}$, we estimate the number of its non-data points, $b_r$, as follow:
\begin{equation}
b_{r} = \left\lfloor\frac{\text{Vol}(r)}{\delta_R^3}\right\rfloor - o_{r}
\end{equation}
where $\text{Vol}(r)$ is a volume of region $r$, $\delta_R$ is a local density of region $R$, and $o_{r}$ is the number of actual points in region $r$.

For any merged region $R$, we compute its local density $\delta_R$ based on nearest neighbor distances $d_i$ for each actual point $p_i$ in $R$ as follows:
\begin{equation}
d_i = \frac{1}{k}\sum_{j=1}^k \|p_i - q_j\|_2,
\end{equation}
\begin{equation}
\delta_R = \frac{1}{o_{R}}\sum_{i=1}^{o_{R}} d_i,
\end{equation}
where $q_j$ represents the $j$-th nearest neighbor of point $p_i$, as shown in Fig. \ref{fig:point_types}(b). The parameter $k$ is typically set to 8 or 16 neighbors, providing a balance between local detail and computational efficiency. Note that using the same $\delta_R$ computed from the merged region for both $r_1$ and $r_2$ ensures consistent density estimation across both regions. As illustrated in Fig. \ref{fig:point_types}(c), this approach naturally adapts to local point cloud characteristics.

While computing $\delta_R$ for each merged region can be computationally intensive, this level of detail is not always necessary. For point clouds with uniform spatial distributions, where points are evenly spaced across the surface, $\delta_R$ remains relatively constant across regions. In such cases, we can optimize performance by computing $\delta_R$ once globally using all points in the cloud, and then apply this single value for non-data points estimation across all regions.

\subsection{Definition of Point Separability}

Let $\hat{\mathcal{P}} = \mathcal{P} \cup \mathcal{B}$ denote our augmented point cloud, where $\mathcal{P}$ represents the set of actual points and $\mathcal{B}$ represents the estimated non-data points. 
Given a point $p \in \hat{\mathcal{P}}$, for any region $R \subset \mathbb{R}^3$, we represent spatial occupancy using a function $f(p)$ defined as:
\begin{equation} \label{eq:occupancy}
f(p) = \begin{cases} 
1 & \text{if } p \in \mathcal{P} \text{ (actual point)} \\
0 & \text{if } p \in \mathcal{B} \text{ (non-data point)}
\end{cases}    
\end{equation}

This binary representation is particularly advantageous when calculating class separability. Since we are dealing with binary values, all object points (1s) and non-data points (0s) contribute equally to the statistical calculations. In this special case, class separability can be computed directly from point counts rather than iterating over individual values, drastically reducing computational overhead. With this in mind, point separability between two adjacent regions, $r_1$ and $r_2$ in a point cloud, $\eta_p$, can be calculated as follows:
\begin{equation} \label{eq:point_separability}
\eta_p = \frac{n_1(\mu_1 - \mu_T)^2 + n_2(\mu_2 - \mu_T)^2}{N\mu_T(1-\mu_T)},
\end{equation}
\begin{equation}
\mu_1 = \frac{o_{1}}{n_1}, \quad \mu_2 = \frac{o_{2}}{n_2}, \quad \mu_T = \frac{o_{1} + o_{2}}{N},
\end{equation}
where $o_{1}$, $o_{2}$ are the number of actual points in ${r_1}$ and ${r_2}$, $b_{1}$, $b_{2}$ are the number of non-data points in ${r_1}$ and ${r_2}$, $n_1 = o_{1} + b_{1}$ and $n_2 = o_{2} + b_{2}$ are the total points in ${r_1}$ and ${r_2}$, and $N = n_1 + n_2$ is the total points in both regions.

An important observation from Eq. \ref{eq:point_separability} is that while we need to know the number of non-data points in each region ($b_1$ and $b_2$), we never use their actual spatial positions in the calculation. This means that in practice, we do not need to explicitly add estimated non-data points to the point cloud, further reducing computational costs by simply estimating the count values for non-data points in each region.

\begin{figure}[t]
    \centering
    \includegraphics[width=5cm]{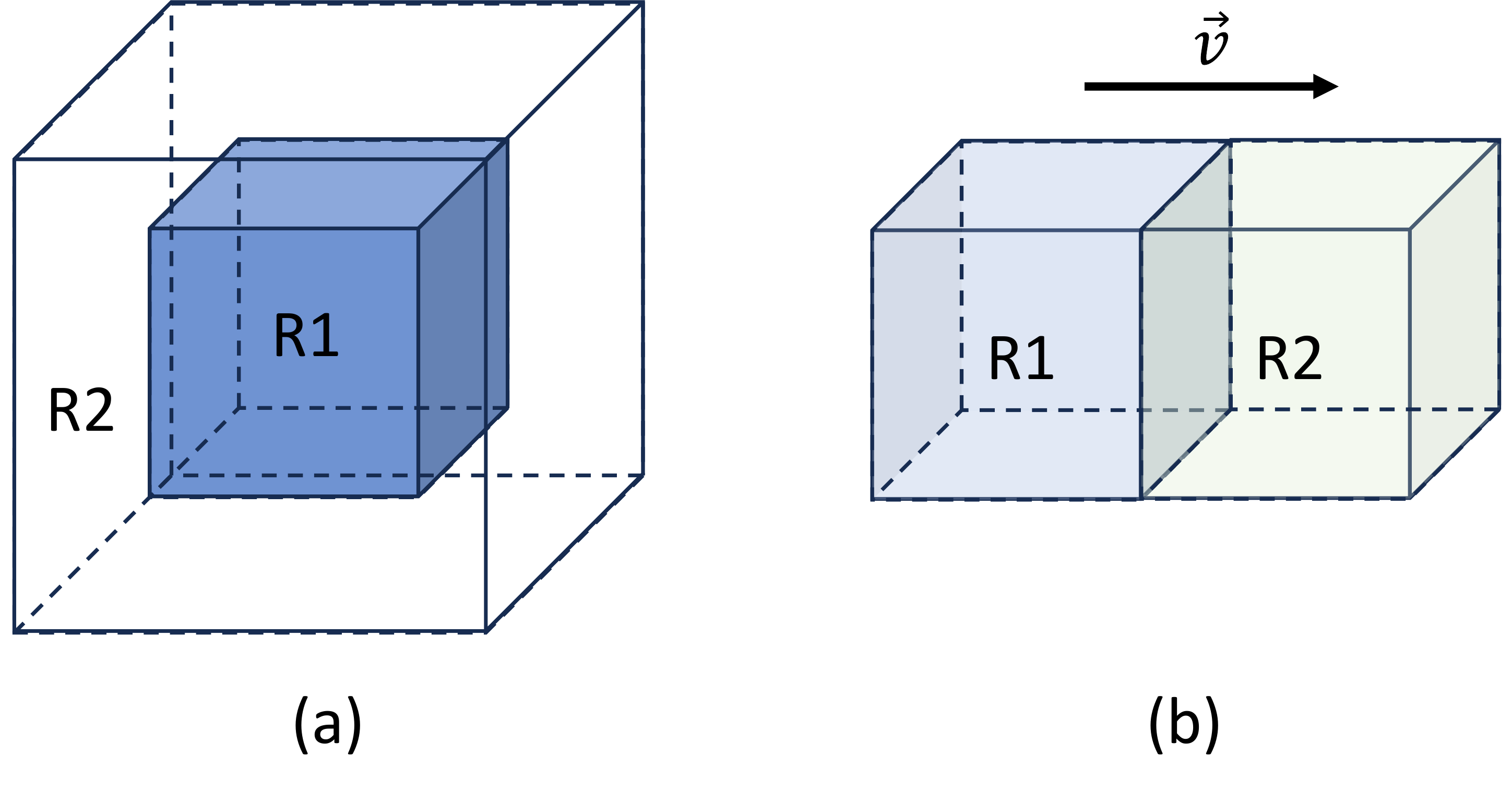}
    \caption{Comparison of 3D separability filter approaches. (a) 3D separability filter using inside/outside cube volumes\cite{Yataka2017FeaturePE}. (b) Our proposed approach that maintains the essence of the original 2D separability filter \cite{fukui1995edge} by comparing adjacent regions, but extends it to 3D space with arbitrary orientation enabled by the point cloud representation.}
    \label{fig:separability_comparison}
\end{figure}

To implement this separability calculation in practice, we must carefully consider how to arrange and orient the comparison regions. Instead of comparing inside and outside cube volumes \cite{Yataka2017FeaturePE}, we maintain the original principle of comparing adjacent regions \cite{fukui1995edge} and extend it to point cloud representation as shown in Fig. \ref{fig:separability_comparison}. This extension naturally enables unrestricted orientation in 3D space, where we can define two adjacent volumes $r_1$ and $r_2$ along any direction vector $\vec{v}$ at spatial point $sp = (x,y,z)$.

\subsection{Weighted Point Separability}
Point clouds often carry multiple attributes beyond spatial coordinates, such as intensity, color, normal vectors, or local density. A point $p$ is characterized by its position and a set of attributes $p = (x, y, z, a^1, a^2, ..., a^d)$ where $(a^1, a^2, ..., a^d)$ represents $d$ different point attributes. To incorporate multiple attributes into our separability calculation, we compute both the point separability $\eta_p$ using Eq. \ref{eq:point_separability} and the separability $\eta_{a^i}$ for each attribute using Eq. \ref{eq:separability}. These separability measures are then combined using a weighted average.

Let $\eta_p$ be the point separability calculated using Eq. \ref{eq:point_separability} for spatial information, and $\eta_{a^i}$ be the separability of the $i$-th attribute calculated using Eq. \ref{eq:separability}. These values form a separability vector $\overline{\eta} = [\eta_p, \eta_{a^1}, \eta_{a^2}, ..., \eta_{a^d}]$. Given a weight vector $\overline{w} = [\overline{w}_1, \overline{w}_2, \dots, \overline{w}_{d+1}]$, where each $\overline{w}_j$ represents the importance of the corresponding separability measure, the weighted separability $\eta_w$ is calculated as:
\begin{equation} \label{eq:weight_separability}
\eta_w = \frac{\sum_{j=1}^{d+1} \overline{w}_j \overline{\eta}_j}{\sum_{j=1}^{d+1} \overline{w}_j}
\end{equation}
where $\overline{w}_j$ is the weight for the $j$-th element in $\overline{\eta}$.

\subsection{Separability Membrane for Surface Reconstruction} 
The separability membrane extends the classical active contour framework to three-dimensional space. While active contours employ energy minimization to deform a 2D contour toward object boundaries, our approach adapts this principle to 3D surface reconstruction through an energy formulation based on point and class separability. 

\begin{figure}[t]
    \centering
    \includegraphics[width=7.0cm]{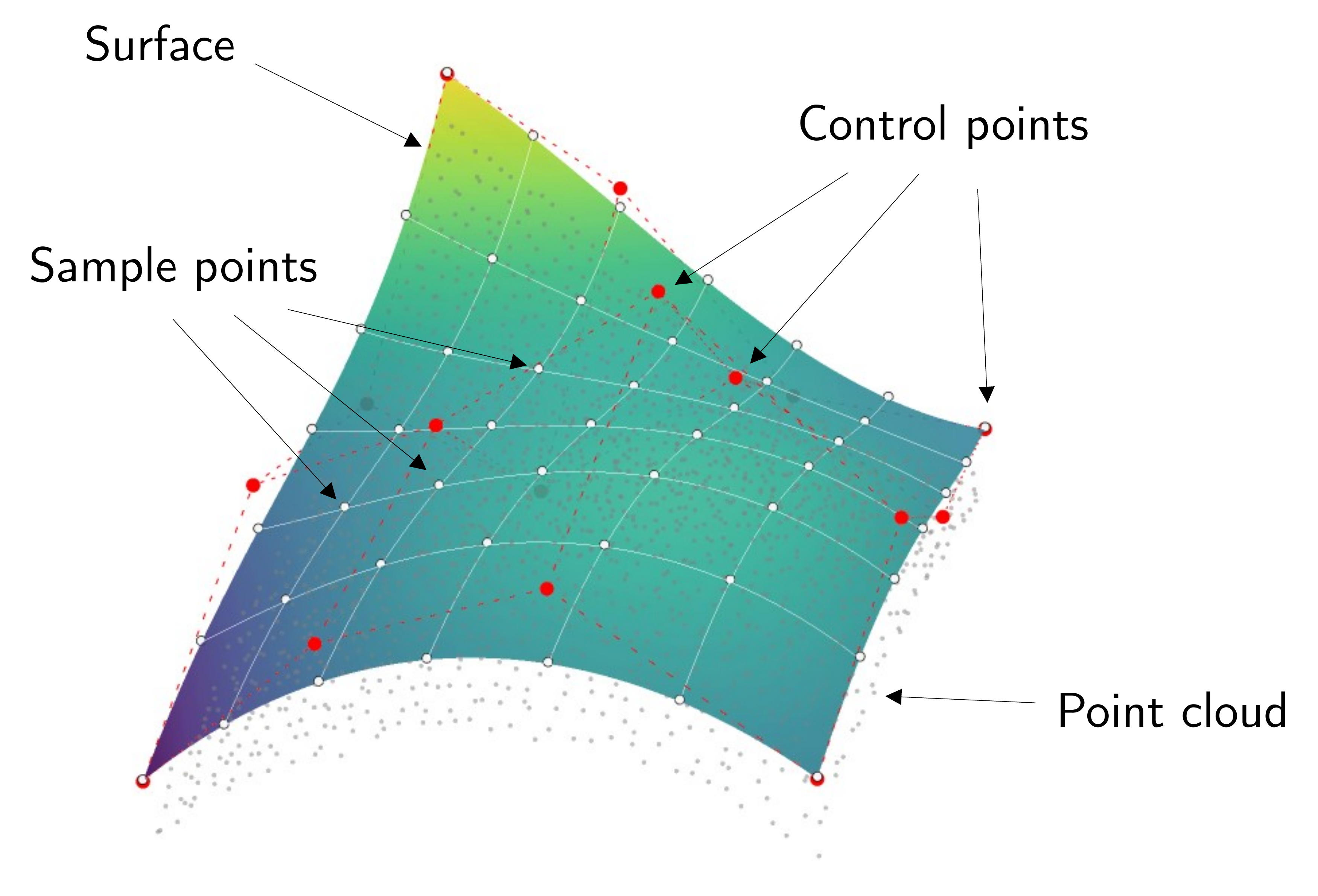}
    \caption{We construct our membrane model using a B-spline surface that are controlled by multiple control points. Red points are control points of the surface. White points are sampled from the surface. Gray points are an input point cloud.}
    \label{fig:b-surface}
\end{figure}

We consider two kinds of energy, internal and external energy, to be optimized to construct our membrane model. 
For the internal energy, we model the membrane with a cubic B-spline surface that is determined by multiple control points and B-spline basis functions\cite{brigger2000b}, as illustrated in Fig. \ref{fig:b-surface} . By using B-spline surface, we do not need to explicitly consider the internal energy of the membrane since the B-spline representation always ensures the minimization of internal energy \cite{cipolla1990dynamic, cipolla1992surface}.  A B-spline surface is calculated by $M \times L$ control points and multiple B-spline basis functions as follows:
\begin{eqnarray}
S(u,v) &=& \sum_{i=0}^M \sum_{j=0}^L N_{i,3}(u) N_{j,3}(v) Q_{i,j},
\label{bspline}
\end{eqnarray}
where $u$ and $v$ are the parameters that determine a sample point on the surface model. $N_{i,3}(u)$ and $N_{j,3}(v)$ are the cubic B-spline basis functions, and $Q_{i,j}$ is the $ij$-th control point \cite{piegl1996nurbs}. 

\begin{figure}[t]
    \centering
    \includegraphics[width=8.0cm]{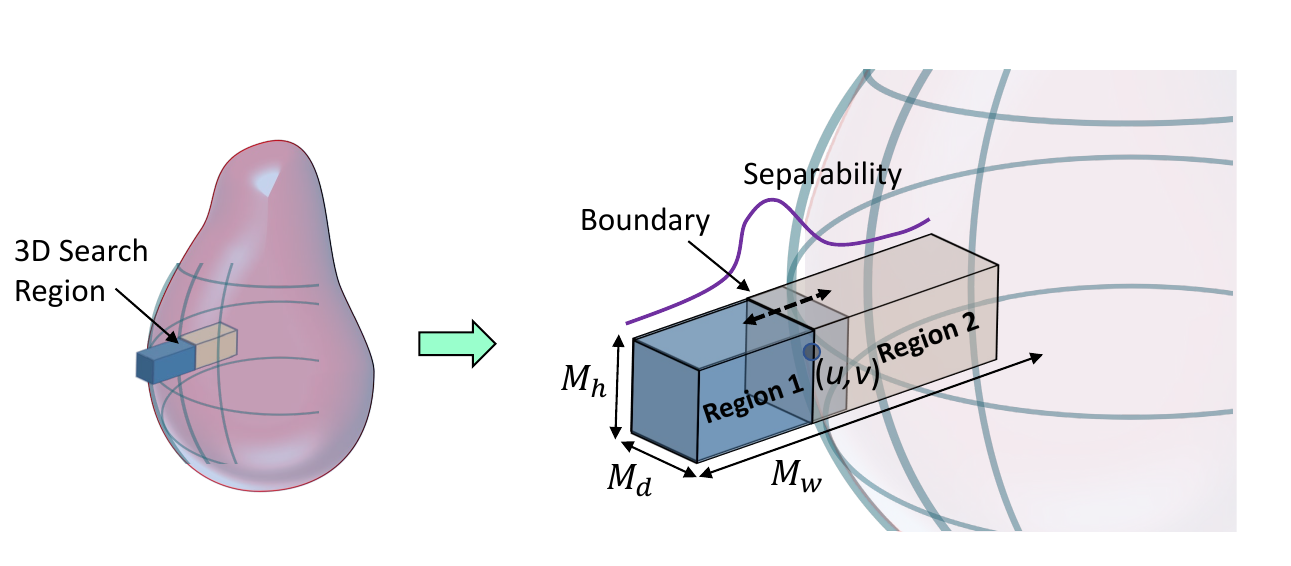}
    \caption{Calculation of external energy at each sample point. A cuboid search region of size $M_d{\times}M_h{\times}M_w$ is set, and the separability is calculated by moving the boundary between regions 1 and 2 within this region.}
    \label{fig:sep_maskshift}
\end{figure}

For the external energy, $E_{ext}$, of the membrane, we introduce the following function defined as follows:
\begin{equation}
    E_{ext} = \int \int E_{sep}(S(u,v))dudv
\end{equation}
\begin{equation}
    E_{sep}(S(u,v)) = -\eta^*(S(u,v))
\end{equation}
where $\eta^*(S(u,v))$ indicates the separability at the sample point $(u,v)$. As shown in Fig. \ref{fig:sep_maskshift}, initially, we set a rectangular cuboid search region with size $M_w$, $M_h$, and $M_d$. Then, we shift the boundary between Region 1 and Region 2 within this search region, changing the volumes of inner and outer regions iteratively to find the position that maximizes separability. The maximum separability obtained is set to $\eta^*(S(u,v))$.

\begin{figure}[t]
    \centering
    \includegraphics[width=8cm]{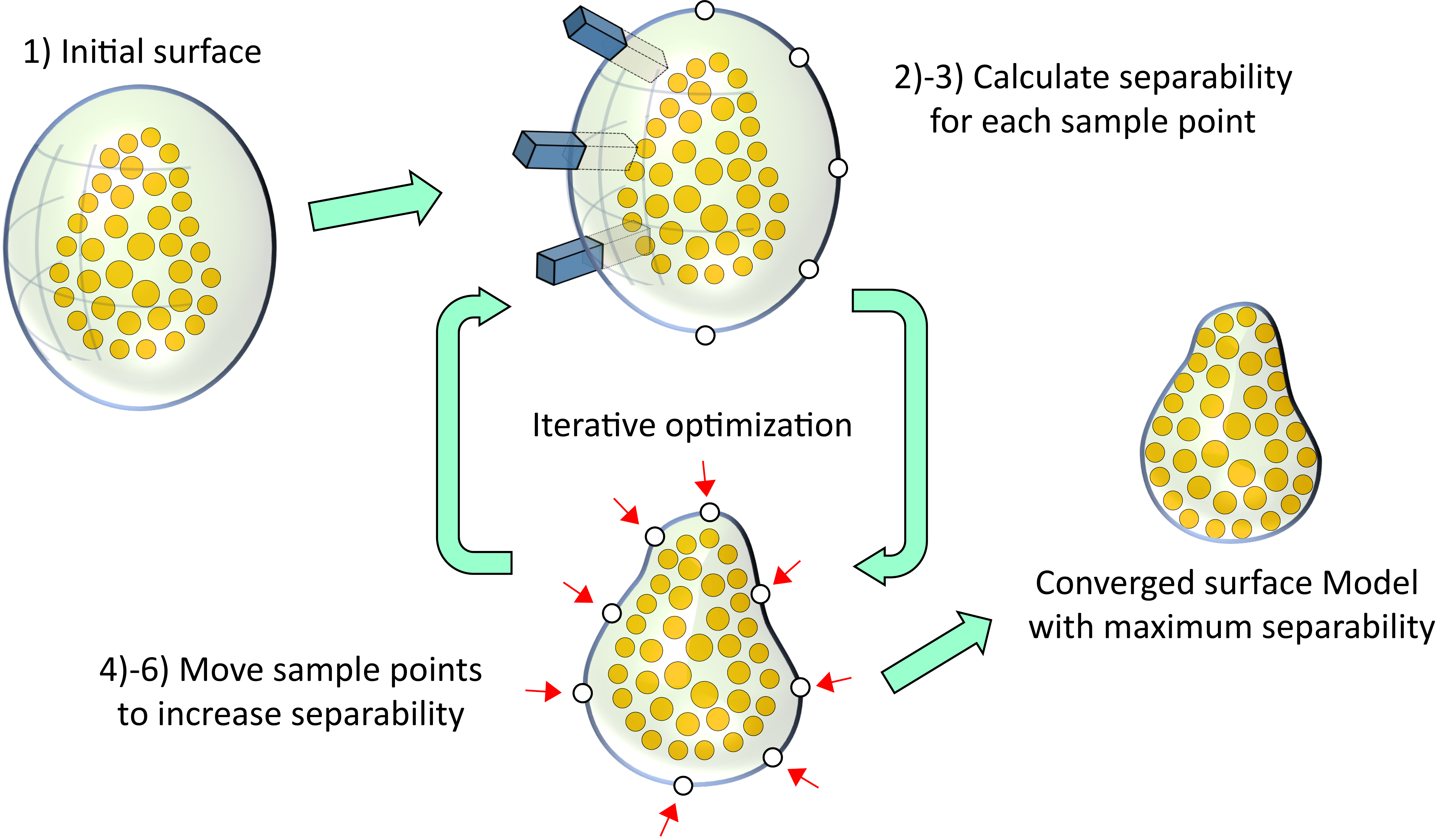}
    \caption{Overview of separability membrane for point cloud surface reconstruction. Given an initial surface, the membrane iteratively calculates the separability at each sample point within a search region, moves the sample points towards maximum separability, and approximates a new B-spline surface. The process repeats until convergence, based on the global separability of all sample points.}
    \label{fig:sep_mem_algorithm}
\end{figure}

We summarize the process of the separability membrane, illustrated in Fig. \ref{fig:sep_mem_algorithm}, as follows:
\begin{enumerate}
\item Set an initial surface model.
\item Sample the current B-spline surface at fixed intervals $(u_{n} \times v_{n})$ to obtain sample points.
\item For each sample point $(u,v)$:
    \begin{itemize}
    \item[a)] Set a rectangular cuboid search region in the direction perpendicular to the surface.
    \item[b)] Calculate the separability $\eta^*(S(u,v))$ using Eq. \ref{eq:point_separability} or Eq. \ref{eq:weight_separability}.
    \item[c)] Move the sample point slightly toward the new point where the separability maximizes in the search region with a shift factor $\beta(<1.0)$.
    \end{itemize}
\item Approximate a new B-spline surface by computing control points that minimize the least squares error to the sample points.
\item Check the stopping condition based on the global separability value that can be calculated as the mean separability of all sample points for the current iteration:
    \begin{itemize}
    \item[a)] If the global separability shows no significant improvement for several consecutive iterations, stop the process.
    \item[b)] Otherwise, return to step 2 and continue iteration.
    \end{itemize}
\end{enumerate}

For the initial surface model placement, we utilize a method inspired by extreme point detection approaches\cite{extremeclick, deepextremecut, octagonmask}. The process identifies six extreme points corresponding to the top-most, bottom-most, front-most, back-most, left-most, and right-most positions of the 3D object. These points can be obtained automatically for well-isolated objects. Following the identification of extreme points, 3D octagon surface are constructed to form the initial surface configuration.

\subsection{Dynamic Membrane Adjustment}
The shape of the active surface model is deformed based on the position of multiple 3D control points of the B-Spline surface and the size of the search region around each control point. While a higher number of control points allows the surface model to express more complex 3D shapes, it also increases computational costs and potentially reduces model stability. Similarly, larger search regions may help in finding the object's boundary but can also make the surface unstable by detecting irrelevant points and increasing computational complexity.

Based on these observations, we initialize the model with a small number of control points and a large search area. The parameters are then adjusted incrementally every $h$ iterations, gradually increasing the number of control points while reducing the search area. While this simple incremental approach can work for objects with known shape characteristics, there is no guarantee it will work for arbitrary point cloud shapes. For instance, complex objects may require different numbers of control points along the u and v parametric directions based on their geometric features. To address these limitations, we propose a dynamic membrane function that automatically adjusts the number of control points in each parametric direction throughout the segmentation process.

Our approach implements an adaptive strategy for control point management. Starting with fewer control points to capture the global 3D shape, it progressively adds more points to represent detailed features as needed. For a surface $S(u,v)$ controlled by $M \times L$ control points, where $M$ and $L$ represent the number of points in each parametric direction, we define rows $R_{i} = \{ Q_{i,1}, Q_{i,2}, ..., Q_{i,N} \}$ and columns $C_{j} = \{ Q_{1,j}, Q_{2,j}, ..., Q_{M,j} \}$. New control points will be added between existing rows and columns based on the average separability of sample points on the surface.

We summarize the function of adding control points as follows:
\begin{enumerate}
\item[(1)] After each iteration of the separability membrane, evaluate the global separability, $\eta_g$, by averaging values of separability from all sample points in the current B-spline surface, $\eta^*(S(u,v))$.
\item[(2)] When the difference between current and previous global separability falls below threshold $g_{min}$, add new $m \times l$ control points to the membrane.
\item[(3)] Continue iteration without adding control points when the difference in global separability between iterations meets or exceeds threshold $g_{min}$.
\end{enumerate}

To complement the dynamic control point adjustment, we also adjust the number of sample points along each parametric direction, $u_n$ and $v_n$, dynamically. The number of divisions in each direction, $div_n$, is calculated as follows:
\begin{equation}
div_n = \max\left(div_{\min}, \alpha(q-1)+1\right)
\end{equation} 
where $div_{\min}$ is the minimum allowed divisions, $q$ is the number of control points in the calculated direction, and $\alpha$ is an adjustable scalar (default $\alpha = 2$) that can be tuned to dataset complexity. This ensures sufficient sampling while maintaining computational efficiency.

%% file: Chapters/4_experiments.tex
\section{Evaluation Experiments}

In this section, we conduct a series of experiments to validate our proposed method. First, we investigate the membrane's ability to express complex shapes using various control point mechanisms. Second, we evaluate our method on a 3D surface extraction task on synthetic noisy image data and a 3D medical image dataset. Finally, we observe our method against irregular 3D data on the point cloud segmentation task.

\begin{figure}[t]
    \centering
    \includegraphics[width=7.0cm]{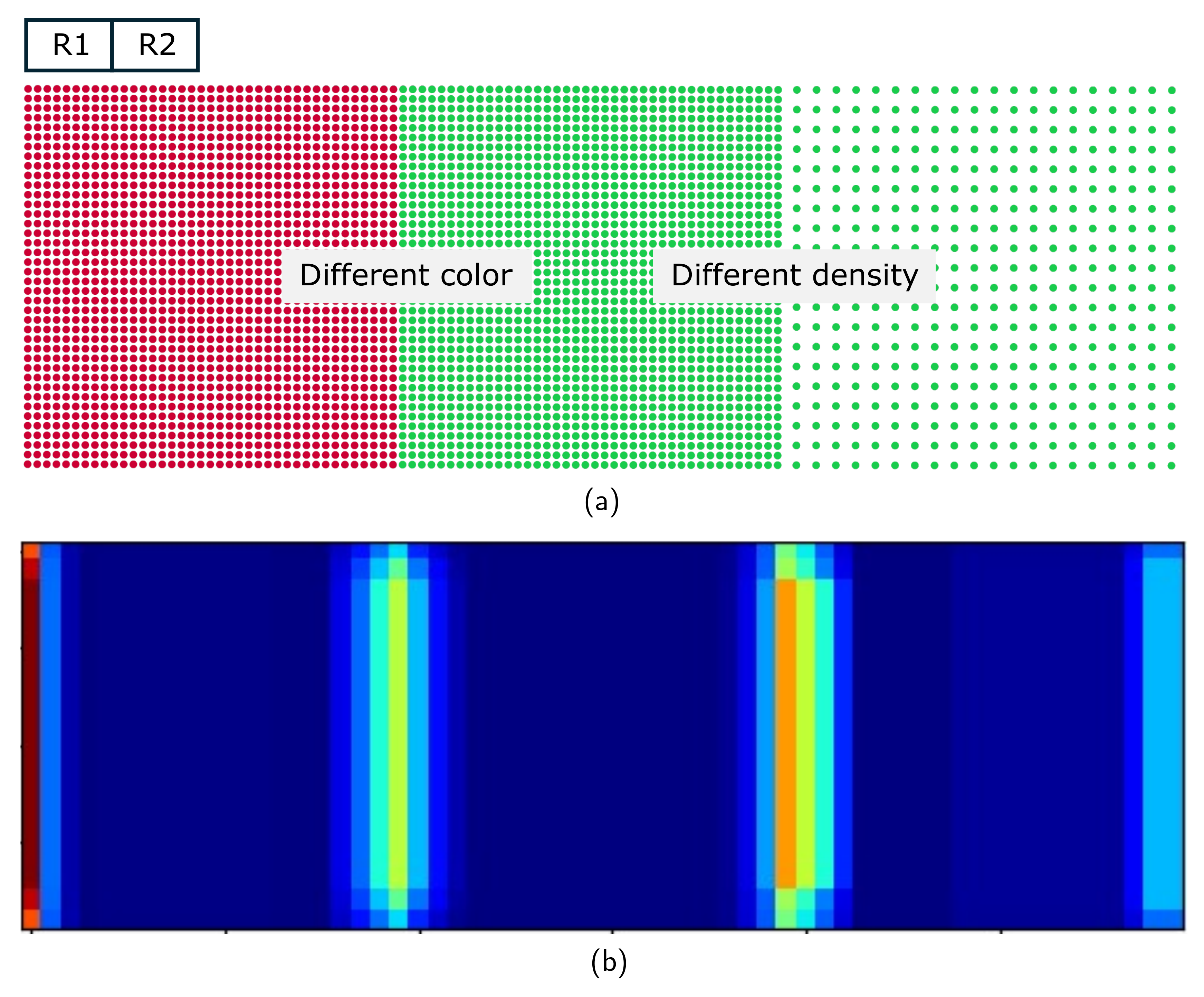}
    \caption{Weighted separability map that consider spatial positions and color features. (a) a point cloud of a plane with different color and density (b) Separability map obtained by doing 3D separability filter.}
    \label{fig:3dsepexp}
\end{figure}

\subsection{Validity of Class Separability for Point Cloud}

To verify the validity of our proposed 3D separability filter for point cloud data, we conducted experiments using both synthetic and real-world datasets. Our evaluation examined the filter's ability to process point cloud data directly and extract meaningful features.

First, we generated a synthetic 2D point cloud data with controlled variations in color and density. The experimental procedure involved sampling points on a 2D grid and applying our filter to compare adjacent regions, as illustrated in Fig. \ref{fig:separability_comparison}(b). The filter computed a separability measure at each grid point, producing a separability map that quantifies the distinction between neighboring regions.

The results demonstrated in Fig. \ref{fig:3dsepexp} confirm that our proposed filter successfully operates directly on point cloud data without requiring conversion to volumetric representation. The separability map shows clear differences in areas where either position or color changes. This multi-attribute sensitivity suggests that our filter can serve as an effective edge detection mechanism for point cloud data, incorporating not only spatial positions but also additional features such as intensity, colors, normal vectors, and density distributions.

\label{sec:exp_res}

\subsection{Robustness to Ambiguous Boundary}

To evaluate the robustness of the separability membrane against ambiguous boundaries, we conducted boundary surface extraction experiments on 3D point cloud data with varying outlier ratios. We compared our method against the Poisson Surface Reconstruction (PSR) algorithm \cite{chan2001active}, a widely used approach for point cloud surface reconstruction. The experimental setup used synthetic point cloud data of a sphere comprising 1000 points. To generate outliers, we created an additional 1000 points positioned at the same locations as the original points, then perturbed them using Gaussian noise with $\sigma$ ratios ranging from 0 to 0.5.

We applied both boundary extraction methods to these point cloud datasets and evaluated their performance using the FScore 1\% metric, which quantifies the overlap between the correct and estimated volumes. For the implementation of the separability membrane, we constructed the surface model using 64 control points ($M{\times}L=8{\times}8$ in Eq.(\ref{bspline})) with a cubic B-spline surface. In addition, we placed 40 sample points for cuboid search areas on each segment surface of the model.

\begin{figure}[t]
    \centering
    \includegraphics[width=8.0cm]{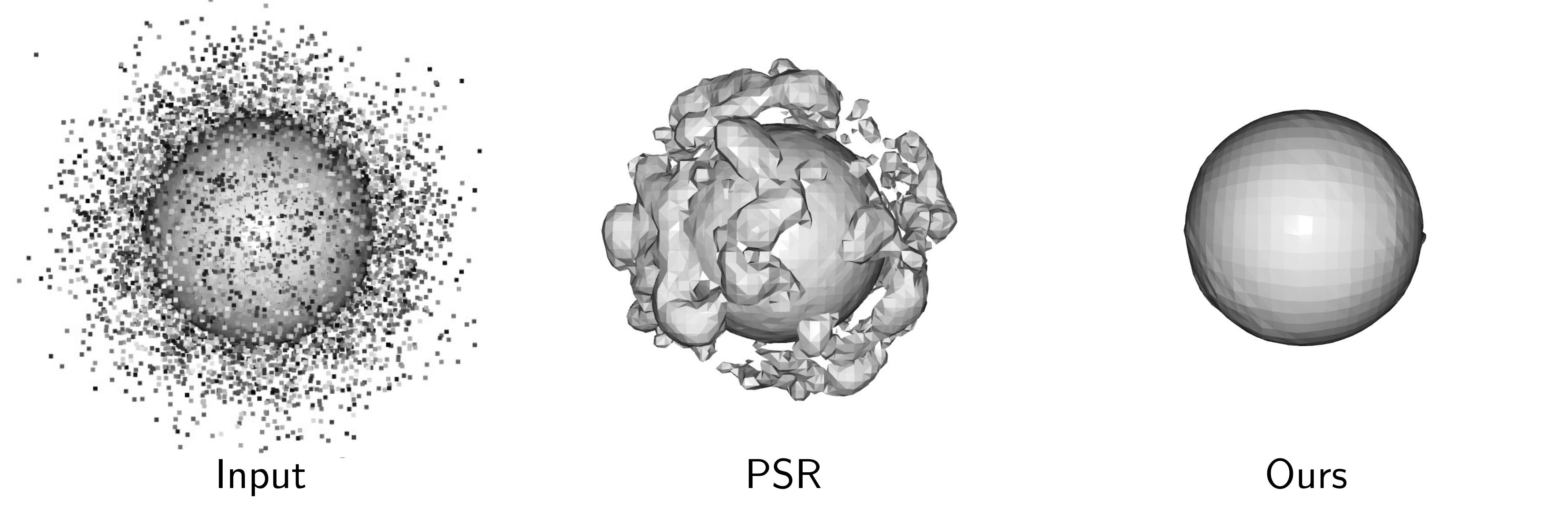}
    \caption{Point cloud surface reconstruction results on point cloud data with outliers ($\sigma=11\%$.)}
    \label{fig:dice_per_sn_ratio}
\end{figure}

\begin{figure}[t]
    \centering
    \includegraphics[width=8.0cm]{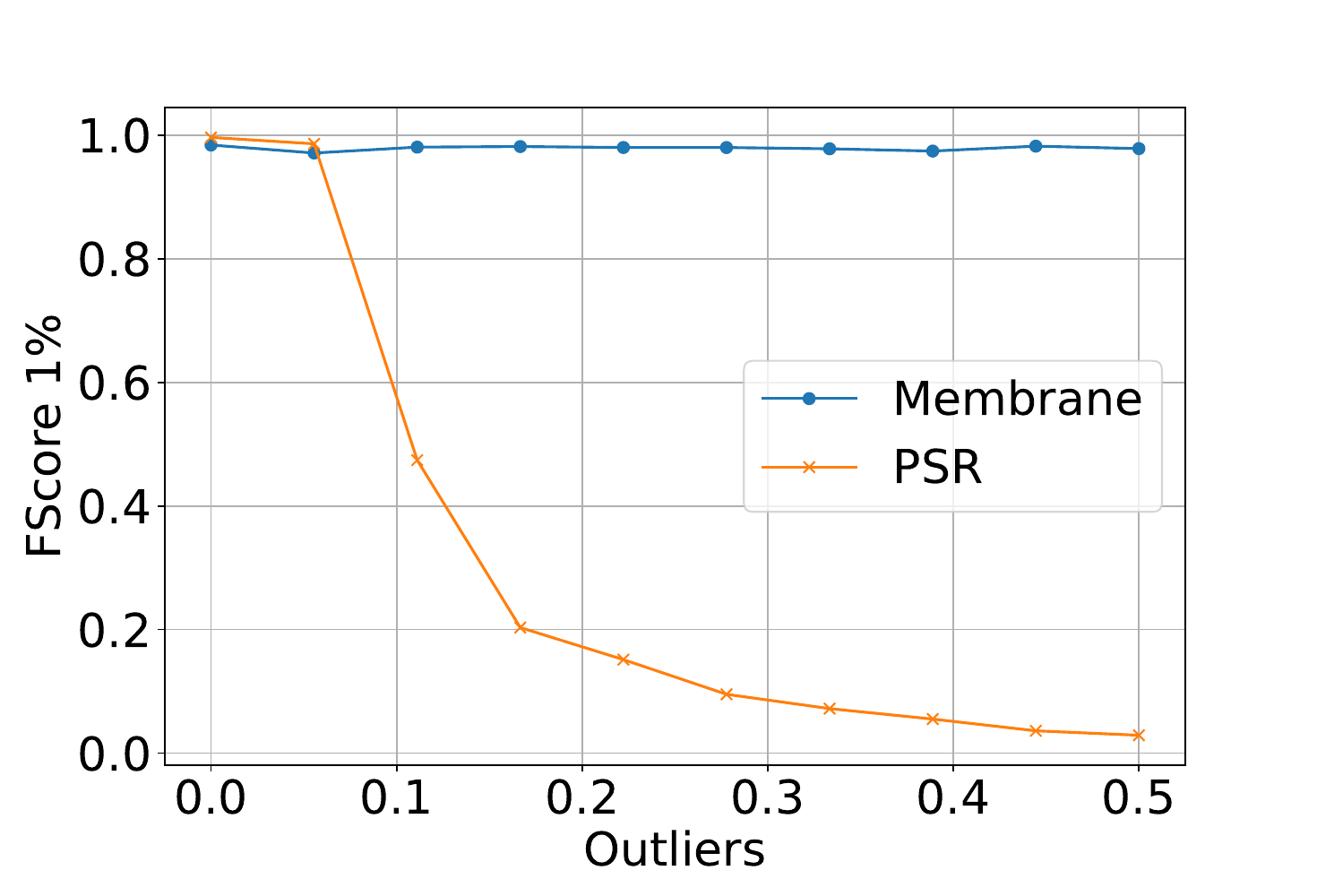}
    \caption{Variation of FScore 1\% per outliers $\sigma$ of our membrane and PSR.}
    \label{fig:noise_cut}
\end{figure}

Fig. \ref{fig:dice_per_sn_ratio} presents the original point cloud with outliers ($\sigma$=0.11) alongside the output meshes generated by the separability membrane and PSR. The results demonstrate that our membrane's surface model effectively converges to the object region, evidenced by the smooth output mesh generated from the active surface. Fig. \ref{fig:noise_cut} compares the FScore 1\% metrics with different outlier intensities. As the $\sigma$ ratio increases, indicating higher outlier levels, the accuracy of PSR decreases significantly. In contrast, the separability membrane maintains relatively stable accuracy despite increasing outlier levels, demonstrating its robustness in handling ambiguous boundaries caused by outliers.

\subsection{Experiments on Dynamic Membrane Adjustment}
We evaluate our membrane's ability to express complex shapes from the 3DNet dataset \cite{3dnet} using dynamic membrane adjustment. The experiment compares three control point mechanisms: static $8 \times 5$, static $40 \times 25$, and an adaptive mechanism that begins with $8 \times 5$ control points and can expand up to $40 \times 25$. We maintain a fixed 3D search region while measuring the Chamfer distance between the surface model and input point cloud across iterations.

\begin{figure}[t]
    \centering
    \includegraphics[width=8cm]{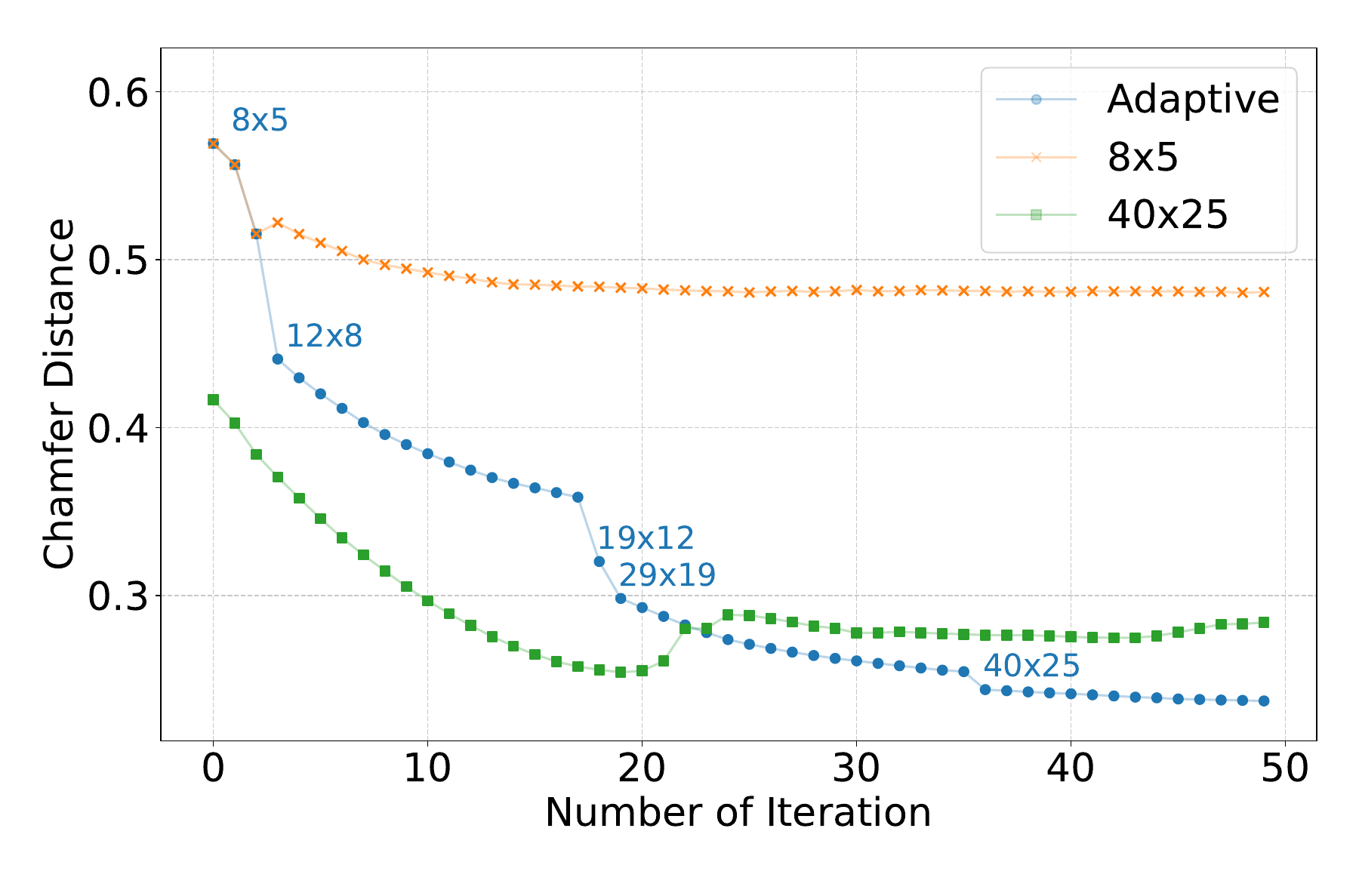}
    \caption{Comparative analysis of Chamfer distance convergence across iterations for static ($8 \times 5$, $40 \times 25$) and adaptive control point mechanisms. Small number of control points struggle to represent complex shape, while too much control points make the membrane unstable and computationally expensive.}
    \label{fig:add_control_exp}
\end{figure}

Fig. \ref{fig:add_control_exp} demonstrates the comparative performance of these mechanisms. It is observed that the static $8 \times 5$ struggle to accurately represent the complexity of the 3D object, while the static $40 \times 25$,, despite its potential for detail, introduces unnecessary computational overhead and membrane instability from the initial iteration. The adaptive mechanism, though initially performing below the static $40 \times 25$, achieves superior results in later iterations while avoiding premature computational burden.

Analysis of Fig. \ref{fig:add_control_exp} reveals the responsive nature of our adaptive approach. In the first three iterations, the adaptive mechanism's performance increases steadily, tracking closely with the static $8 \times 5$. When the average separability begins to decrease at the fourth iteration, our approach automatically adds more control points to maintain performance, while the static $8 \times 5$ continues to degrade. This observation confirms that the mechanism successfully detects and responds to decreasing separability by adjusting the surface model's rigidity through additional control points. While the adaptive approach may requires additional iterations to converge, it achieves both superior computational efficiency (running two times faster than the static $40 \times 25$) and better accuracy in expressing complex shapes. These results underscore how responsive adaptability in control point mechanisms enhances our separability membrane's ability to capture 3D object shapes.

\subsection{Experiments on Surface Reconstruction}

\begin{figure*}[tb]
    \centering
    \includegraphics[width=\textwidth]{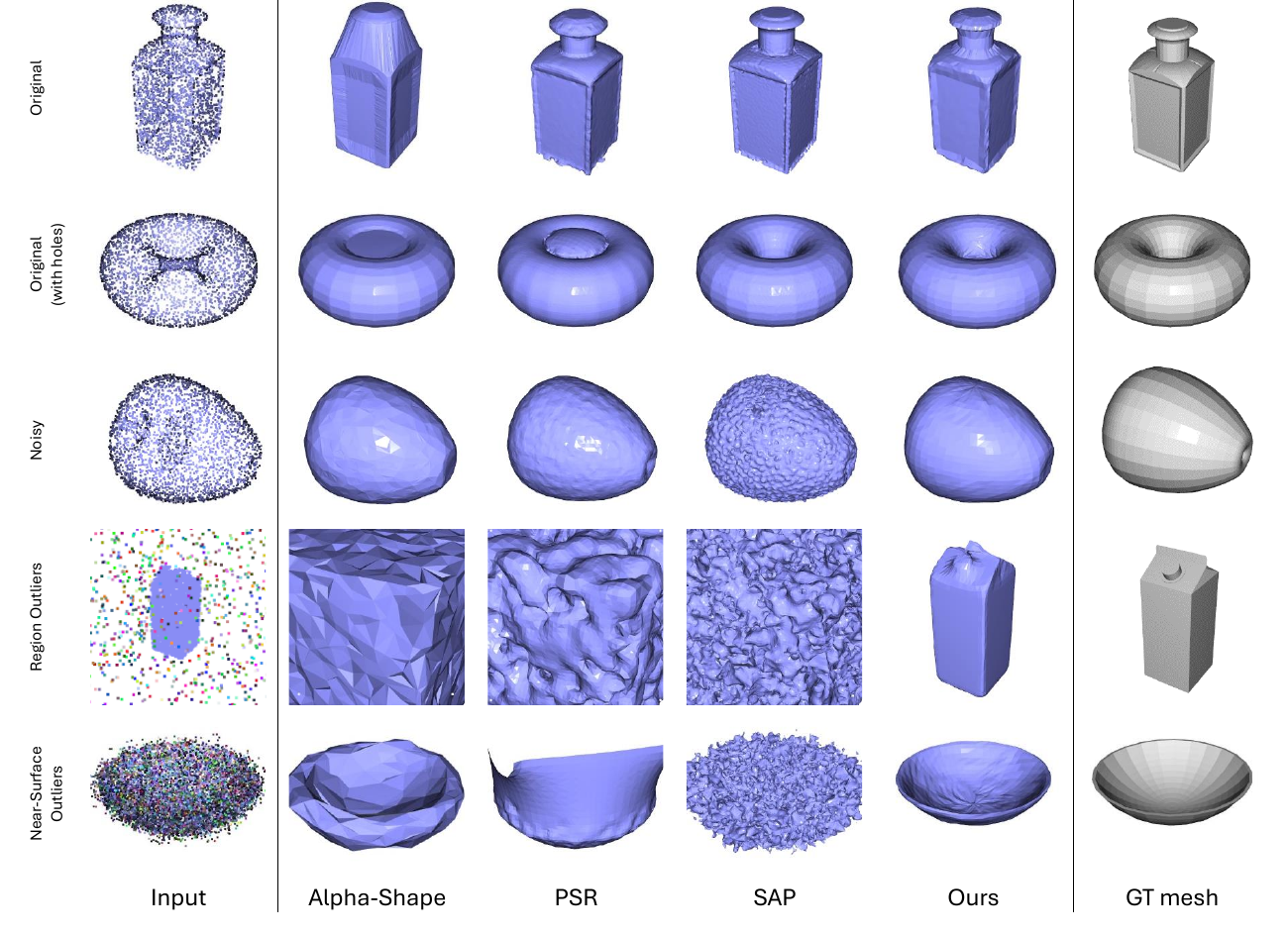}
    \caption{Qualitative comparison on 3DNet point cloud data on various experiments. Input point clouds are downsampled for visualization.}
    \label{fig:pcd_results}
\end{figure*}

We conducted experiments on surface reconstruction using unorganized point cloud data from the 3DNet dataset \cite{3dnet} to explore the effectiveness of our method under various conditions. Five object meshes represented as point cloud data were selected as examples to demonstrate the robustness of our method. In this experiment, we limit our membrane to only considering a spatial position to ensure a fair comparison with existing reconstruction methods. Fig. \ref{fig:pcd_results} shows the input point clouds and their qualitative comparison under various conditions.

\begin{table*}[tb]
\caption{Quantitative comparison on surface reconstruction task on five point-cloud data from 3DNet dataset.}
\label{tab:exp_result}
    \centering
    \begin{tabular}{l|l|c|c|c|c}
    Dataset &Method &Chamfer $(\downarrow)$ &F-Score $(\uparrow)$ &Normal C. $(\uparrow)$ &Time (s)\\
    \hline
    \hline
Original & Alpha-shape\cite{Edelsbrunner1983} & 0.014 & 0.883 & 0.884 & 3.660 \\
 & PSR\cite{Kazhdan2006} & 0.015 & 0.876 & 0.896 & \textbf{0.151} \\
 & SAP\cite{Peng2021SAP} & \textbf{0.009} & \textbf{0.951} & \textbf{0.900} & 80.212 \\
 & \textbf{Ours} & 0.011 & 0.936 & 0.899 & 9.259 \\
 \hline
Noise=0.005 & Alpha-shape\cite{Edelsbrunner1983} & 0.020 & 0.816 & 0.892 & 0.462 \\
 & PSR\cite{Kazhdan2006} & 0.015 & 0.874 & 0.889 & \textbf{0.156} \\
 & SAP\cite{Peng2021SAP} & \textbf{0.010} & \textbf{0.936} & 0.788 & 94.960 \\
 & \textbf{Ours} & 0.012 & 0.915 & \textbf{0.897} & 9.427 \\
 \hline
Region Outliers & Alpha-shape\cite{Edelsbrunner1983} & 1.296 & 0 & 0.638 & 0.675 \\
 & PSR\cite{Kazhdan2006} & 0.546 & 0.037 & 0.748 & \textbf{0.326} \\
 & SAP\cite{Peng2021SAP} & 0.016 & 0.839 & 0.854 & 258.520 \\
 & \textbf{Ours} & \textbf{0.011} & \textbf{0.915} & \textbf{0.893} & 22.769 \\
 \hline
Surface Outliers & Alpha-shape\cite{Edelsbrunner1983} & 0.137 & 0 & 0.822 & 1.169 \\
 & PSR\cite{Kazhdan2006} & 0.025 & 0.831 & 0.850 & \textbf{0.179} \\
 & SAP\cite{Peng2021SAP} & 0.037 & 0.517 & 0.823 & 143.060 \\
 & \textbf{Ours} & \textbf{0.012} & \textbf{0.912} & \textbf{0.889} & 19.076 \\
 \hline
Overall & Alpha-shape\cite{Edelsbrunner1983} & 0.367 & 0.425 & 0.809 & 1.492 \\
 & PSR\cite{Kazhdan2006} & 0.150 & 0.654 & 0.846 & \textbf{0.203} \\
 & SAP\cite{Peng2021SAP} & 0.018 & 0.811 & 0.841 & 144.188 \\
 & \textbf{Ours} & \textbf{0.011} & \textbf{0.920} & \textbf{0.895} & 15.133
    \end{tabular}
    
\end{table*}

\begin{table*}[tb]
\caption{Ablation study on point attributes integration using 3DNet Dataset with varying outliers.}
    \centering
    \begin{tabular}{c|c|c||ccc|ccc|ccc}
    \multicolumn{3}{c||}{Attributes} & \multicolumn{3}{c|}{(a) No Outliers} & \multicolumn{3}{c|}{(b) Outliers=0.01} & \multicolumn{3}{c}{(c) Outliers=0.025} \\
    \cline{1-12}
    Spatial & Intensity & Density 
    & \multicolumn{1}{c}{Chamfer} & \multicolumn{1}{c}{F-Score} & \multicolumn{1}{c|}{Time (s)} 
    & \multicolumn{1}{c}{Chamfer} & \multicolumn{1}{c}{F-Score} & \multicolumn{1}{c|}{Time (s)} 
    & \multicolumn{1}{c}{Chamfer} & \multicolumn{1}{c}{F-Score} & \multicolumn{1}{c}{Time (s)}  \\
    \hline
    \hline

    \checkmark &  &  & \textbf{0.01087} & \textbf{0.937} & \textbf{14.730} & 0.01207 & 0.915 & \textbf{28.932} & 0.01206 & 0.909 & \textbf{30.501} \\
     & \checkmark &  & 0.01088 & 0.936 & 22.498 & 0.01164 & 0.919 & 42.891 & 0.01119 & 0.923 & 43.991 \\
     &  & \checkmark & 0.01258 & 0.903 & 24.156 & 0.01155 & 0.920 & 46.065 & 0.01176 & 0.918 & 48.284 \\
    \checkmark & \checkmark &  & 0.01089 & 0.936 & 23.992 & 0.01178 & 0.919 & 44.111 & 0.01146 & 0.918 & 44.663 \\
    \checkmark &  & \checkmark & 0.01159 & 0.925 & 25.768 & 0.01157 & 0.921 & 46.134 & 0.01146 & 0.920 & 47.704 \\
     & \checkmark & \checkmark & 0.01155 & 0.924 & 32.933 & \textbf{0.01123} & \textbf{0.927} & 59.520 & \textbf{0.01125} & \textbf{0.926} & 60.846 \\
    \checkmark & \checkmark & \checkmark & 0.01089 & 0.934 & 35.244 & 0.01157 & 0.922 & 64.536 & \textbf{0.01125} & 0.925 & 64.042 \\

    \end{tabular}
    \label{tab:color_exp}
\end{table*}

To evaluate the performance both quantitatively and qualitatively, we compared our optimization-based approach with three representative surface reconstruction methods: Alpha-shape \cite{Edelsbrunner1983}, Poisson Surface Reconstruction (PSR) \cite{Kazhdan2006}, and Optimization-based Shape-as-points (SAP) \cite{Peng2021SAP}. We consider Chamfer distance, normal consistency, and F-Score with the default threshold of 1\% for this evaluation. All experiments were conducted on a desktop computer equipped with an Intel Core i7-13700F processor (24 cores) and an NVIDIA GeForce RTX 4070 Ti GPU.

From Table \ref{tab:exp_result}, each method demonstrates distinct strengths and limitations across different scenarios. Alpha-shape, while computationally efficient, shows significant performance degradation in the presence of outliers, particularly struggling with region and surface outliers where it fails to produce meaningful reconstructions. This indicates its high sensitivity to noise and outliers, making it less suitable for real-world applications where data quality cannot be guaranteed.

PSR demonstrates consistent and balanced performance across most scenarios, maintaining reasonable scores in all metrics with the fastest processing times, running approximately 60 times faster than SAP and 40 times faster than our method. However, PSR's performance is limited by its reliance on normal estimation, particularly struggling with thin surfaces and complex geometries where normal estimation becomes unreliable.

SAP achieves the best performance on clean data and slightly degrades its performance with moderate noise. It is able to handle complex geometries, objects with holes, and thin surfaces through optimizing points and normals guided by the Chamfer distance. However, its heavy reliance on Chamfer distance makes it vulnerable to outliers, with F-Score dropping by nearly 98\% for region outliers and 55\% for surface outliers compared to clean data. Additionally, SAP's computational cost is substantial, requiring roughly 8-9 times longer processing time than our method, which may limit the scalability and its practical applications.

Our proposed method demonstrates robust performance across all scenarios, especially in challenging cases with outliers. In the overall comparison, it achieves the best scores across all metrics while maintaining reasonable processing times. Like SAP, our method does not require any normal estimation and is thus more robust to noise. Using normal vectors from a contracting membrane is proven to be effective in handling objects with holes and thin surfaces, such as donuts and bowls, while maintaining better robustness to noise and outliers. 

The qualitative comparison in Fig. \ref{fig:pcd_results} further supports these findings. On clean data, while both SAP and our method achieve high-quality surface reconstruction, Alpha-shape struggles with objects containing holes, and PSR shows limitations when handling thin surfaces. When tested against region outliers, existing methods fail to produce meaningful reconstructions. In the case of surface outliers, PSR demonstrates some degree of robustness, while both Alpha-shape and SAP show significant deterioration in performance, highlighting their sensitivity to this type of noise. In contrast, our method maintains consistent reconstruction quality across all conditions, producing results nearly identical to those obtained with clean data.

\subsection{Ablation Study on Point Attributes}
To further evaluate the effectiveness of our method, we conducted additional experiments incorporating different point attributes: spatial coordinates (XYZ), intensity (derived from RGB values), and local density (computed as the number of neighboring points within a fixed radius). We evaluated seven variants by selectively using these attributes. For testing robustness, we created outliers by duplicating the original point cloud and applying Gaussian noise with different standard deviations ($\sigma = 0.01$ and $\sigma = 0.025$) to these duplicated points. To create distinct appearance characteristics, we colored the original object points white while assigning random darker colors to the outlier points.

The results in Table \ref{tab:color_exp} demonstrate interesting trade-offs between different attribute combinations. The spatial-only variant achieves the best performance with clean data but degrades as noise increases, showing the limitation of purely geometric information in handling outliers. Using only density information results in worse overall performance, yet shows consistent behavior across different noise levels, suggesting its role as a stable but less discriminative feature. The intensity-only variant matches spatial-only performance in clean data and maintains better performance with outliers, benefiting from the distinct color difference between surface points and outliers. The combined approach successfully leverages these complementary signals, spatial coordinates for basic structure, intensity for color-based discrimination, and density for neighborhood consistency, achieving the best overall performance while maintaining practical computational cost due to simple attribute calculations. This validates our weighted separability formulation's effectiveness in combining multiple point attributes to enhance the reconstruction quality of the proposed method.

%% file: Chapters/5_conclusion.tex
\section{Conclusion}
In this paper, we proposed the Separability Membrane, a novel 3D active contour approach for point cloud surface reconstruction. This method defines the surface of a 3D object as the boundary that maximizes the separability between its inner and outer regions using Fisher's ratio. The membrane gradually deforms to identify the exact boundary while maximizing separability between the two volume regions, without requiring training data or conversion to volumetric representation. Despite its simple formulation, the Separability Membrane demonstrates robust performance, accurately reconstructing surfaces even in the presence of noise, outliers, and complex geometries. Through comprehensive evaluations on synthetic data and the 3DNet dataset, we validated the method's effectiveness under diverse conditions. Furthermore, incorporating additional point attributes has been shown to improve reconstruction quality, particularly under challenging conditions. This versatile framework offers significant potential for applications in computer vision and robotics, with promising extensions for higher dimensional data (4D) analysis and real-time point cloud processing in robot vision systems.

%% file: main.bbl
\begin{thebibliography}{10}
\providecommand{\url}[1]{#1}
\csname url@samestyle\endcsname
\providecommand{\newblock}{\relax}
\providecommand{\bibinfo}[2]{#2}
\providecommand{\BIBentrySTDinterwordspacing}{\spaceskip=0pt\relax}
\providecommand{\BIBentryALTinterwordstretchfactor}{4}
\providecommand{\BIBentryALTinterwordspacing}{\spaceskip=\fontdimen2\font plus
\BIBentryALTinterwordstretchfactor\fontdimen3\font minus \fontdimen4\font\relax}
\providecommand{\BIBforeignlanguage}[2]{{%
\expandafter\ifx\csname l@#1\endcsname\relax
\typeout{** WARNING: IEEEtran.bst: No hyphenation pattern has been}%
\typeout{** loaded for the language `#1'. Using the pattern for}%
\typeout{** the default language instead.}%
\else
\language=\csname l@#1\endcsname
\fi
#2}}
\providecommand{\BIBdecl}{\relax}
\BIBdecl

\bibitem{Alkadri2022Investigating}
M.~Alkadri, S.~Alam, H.~Santosa, A.~Yudono, and S.~Beselly, ``Investigating surface fractures and materials behavior of cultural heritage buildings based on the attribute information of point clouds stored in the tls dataset,'' \emph{Remote. Sens.}, vol.~14, p. 410, 2022.

\bibitem{Moyano2021Operability}
J.~Moyano, J.~E. Nieto-Julián, L.~M. Lenin, and S.~Bruno, ``Operability of point cloud data in an architectural heritage information model,'' \emph{International Journal of Architectural Heritage}, vol.~16, pp. 1588 -- 1607, 2021.

\bibitem{Fryskowska2018A}
A.~Fryskowska and J.~Stachelek, ``A no-reference method of geometric content quality analysis of 3d models generated from laser scanning point clouds for hbim,'' \emph{Journal of Cultural Heritage}, 2018.

\bibitem{defect_detection_multi}
M.~Asad, W.~Azeem, H.~Jiang, H.~T. Mustafa, J.~Yang, and W.~Liu, ``2m3df: Advancing 3d industrial defect detection with multi perspective multimodal fusion network,'' \emph{IEEE Transactions on Circuits and Systems for Video Technology}, pp. 1--1, 2025.

\bibitem{Liu2023Near}
Z.~Liu, D.~Kim, S.~Lee, L.~Zhou, X.~An, and M.~Liu, ``Near real-time 3d reconstruction and quality 3d point cloud for time-critical construction monitoring,'' \emph{Buildings}, 2023.

\bibitem{Wang2019Computational}
Q.~Wang, Y.~Tan, and Z.~Mei, ``Computational methods of acquisition and processing of 3d point cloud data for construction applications,'' \emph{Archives of Computational Methods in Engineering}, vol.~27, pp. 479 -- 499, 2019.

\bibitem{Wang2019Applications}
Q.~Wang and M.~koo Kim, ``Applications of 3d point cloud data in the construction industry: A fifteen-year review from 2004 to 2018,'' \emph{Adv. Eng. Informatics}, vol.~39, pp. 306--319, 2019.

\bibitem{action_recognition}
G.~T. Papadopoulos and P.~Daras, ``Human action recognition using 3d reconstruction data,'' \emph{IEEE Transactions on Circuits and Systems for Video Technology}, vol.~28, no.~8, pp. 1807--1823, 2018.

\bibitem{quadratic}
T.~Sun, G.~Liu, R.~Li, S.~Liu, S.~Zhu, and B.~Zeng, ``Quadratic terms based point-to-surface 3d representation for deep learning of point cloud,'' \emph{IEEE Transactions on Circuits and Systems for Video Technology}, vol.~32, no.~5, pp. 2705--2718, 2022.

\bibitem{Edelsbrunner1983}
H.~Edelsbrunner, D.~Kirkpatrick, and R.~Seidel, ``On the shape of a set of points in the plane,'' \emph{IEEE Transactions on Information Theory}, vol.~29, no.~4, pp. 551--559, 1983.

\bibitem{Kazhdan2006}
M.~Kazhdan, M.~Bolitho, and H.~Hoppe, ``Poisson surface reconstruction,'' in \emph{Proceedings of the Fourth Eurographics Symposium on Geometry Processing}, ser. SGP '06.\hskip 1em plus 0.5em minus 0.4em\relax Goslar, DEU: Eurographics Association, 2006, p. 61–70.

\bibitem{point2mesh}
\BIBentryALTinterwordspacing
R.~Hanocka, G.~Metzer, R.~Giryes, and D.~Cohen-Or, ``Point2mesh: a self-prior for deformable meshes,'' \emph{ACM Trans. Graph.}, vol.~39, no.~4, aug 2020. [Online]. Available: \url{https://doi.org/10.1145/3386569.3392415}
\BIBentrySTDinterwordspacing

\bibitem{parsenet}
G.~Sharma, D.~Liu, S.~Maji, E.~Kalogerakis, S.~Chaudhuri, and R.~M{\v{e}}ch, ``Parsenet: A parametric surface fitting network for 3d point clouds,'' in \emph{Computer Vision -- ECCV 2020}, A.~Vedaldi, H.~Bischof, T.~Brox, and J.-M. Frahm, Eds.\hskip 1em plus 0.5em minus 0.4em\relax Cham: Springer International Publishing, 2020, pp. 261--276.

\bibitem{Peng2021SAP}
S.~Peng, C.~M. Jiang, Y.~Liao, M.~Niemeyer, M.~Pollefeys, and A.~Geiger, ``Shape as points: A differentiable poisson solver,'' in \emph{Advances in Neural Information Processing Systems (NeurIPS)}, 2021.

\bibitem{bpnet}
\BIBentryALTinterwordspacing
R.~Fu, C.~Wen, Q.~Li, X.~Xiao, and P.~Alliez, ``Bpnet: Bézier primitive segmentation on 3d point clouds,'' in \emph{Proceedings of the Thirty-Second International Joint Conference on Artificial Intelligence, {IJCAI-23}}, E.~Elkind, Ed.\hskip 1em plus 0.5em minus 0.4em\relax International Joint Conferences on Artificial Intelligence Organization, 8 2023, pp. 754--762, main Track. [Online]. Available: \url{https://doi.org/10.24963/ijcai.2023/84}
\BIBentrySTDinterwordspacing

\bibitem{bezier}
R.~Fu, Q.~Li, C.~Wen, N.~An, and F.~Tang, ``A novel framework for learning bézier decomposition from 3d point clouds,'' \emph{IEEE Transactions on Circuits and Systems for Video Technology}, pp. 1--1, 2024.

\bibitem{kass1988snakes}
M.~Kass, A.~Witkin, and D.~Terzopoulos, ``Snakes: Active contour models,'' \emph{International journal of computer vision}, vol.~1, no.~4, pp. 321--331, 1988.

\bibitem{caselles1995geodesic}
V.~Caselles, R.~Kimmel, and G.~Sapiro, ``Geodesic active contours,'' in \emph{Proceedings of IEEE international conference on computer vision}, 1995, pp. 694--699.

\bibitem{chan2001active}
T.~F. Chan and L.~A. Vese, ``Active contours without edges,'' \emph{IEEE Transactions on image processing}, vol.~10, no.~2, pp. 266--277, 2001.

\bibitem{voxnet2015}
D.~Maturana and S.~Scherer, ``Voxnet: A 3d convolutional neural network for real-time object recognition,'' in \emph{2015 IEEE/RSJ International Conference on Intelligent Robots and Systems (IROS)}, 2015, pp. 922--928.

\bibitem{Yataka2017FeaturePE}
\BIBentryALTinterwordspacing
R.~Yataka, L.~S. de~Souza, and K.~Fukui, ``Feature point extraction using 3d separability filter for finger shape recognition,'' 2017. [Online]. Available: \url{https://api.semanticscholar.org/CorpusID:221817346}
\BIBentrySTDinterwordspacing

\bibitem{pointnet}
R.~Q. Charles, H.~Su, M.~Kaichun, and L.~J. Guibas, ``Pointnet: Deep learning on point sets for 3d classification and segmentation,'' in \emph{IEEE Conference on Computer Vision and Pattern Recognition}, 2017, pp. 77--85.

\bibitem{voxelization}
\BIBentryALTinterwordspacing
D.~Cohen-Or and A.~Kaufman, ``Fundamentals of surface voxelization,'' \emph{Graphical Models and Image Processing}, vol.~57, no.~6, pp. 453--461, 1995. [Online]. Available: \url{https://www.sciencedirect.com/science/article/pii/S1077316985710398}
\BIBentrySTDinterwordspacing

\bibitem{cipolla1990dynamic}
R.~Cipolla and A.~Blake, ``The dynamic analysis of apparent contours,'' in \emph{Proceedings of IEEE international conference on computer vision}, 1990, pp. 616--617.

\bibitem{wakasugi2004robust}
T.~Wakasugi, M.~Nishiura, and K.~Fukui, ``Robust lip contour extraction using separability of multi-dimensional distributions,'' in \emph{Proceedings in Sixth IEEE International Conference on Automatic Face and Gesture Recognition}, 2004, pp. 415--420.

\bibitem{cipolla1992surface}
R.~Cipolla and A.~Blake, ``Surface shape from the deformation of apparent contours,'' \emph{International journal of computer vision}, vol.~9, pp. 83--112, 1992.

\bibitem{3dnet}
W.~Wohlkinger, A.~{Aldoma Buchaca}, R.~Rusu, and M.~Vincze, ``{3DNet: Large-Scale Object Class Recognition from CAD Models},'' in \emph{IEEE International Conference on Robotics and Automation (ICRA)}, 2012.

\bibitem{Bernardini1999}
F.~Bernardini, J.~Mittleman, H.~Rushmeier, C.~Silva, and G.~Taubin, ``The ball-pivoting algorithm for surface reconstruction,'' \emph{IEEE Transactions on Visualization and Computer Graphics}, vol.~5, no.~4, pp. 349--359, 1999.

\bibitem{mescheder2019occupancy}
L.~Mescheder, M.~Oechsle, M.~Niemeyer, S.~Nowozin, and A.~Geiger, ``Occupancy networks: Learning 3d reconstruction in function space,'' in \emph{Proceedings IEEE Conf. on Computer Vision and Pattern Recognition (CVPR)}, 2019.

\bibitem{abstractionTulsiani17}
S.~Tulsiani, H.~Su, L.~J. Guibas, A.~A. Efros, and J.~Malik, ``Learning shape abstractions by assembling volumetric primitives,'' in \emph{Computer Vision and Pattern Regognition (CVPR)}, 2017.

\bibitem{Li2019SPFN}
L.~Li, M.~Sung, A.~Dubrovina, L.~Yi, and L.~J. Guibas, ``Supervised fitting of geometric primitives to 3d point clouds,'' in \emph{2019 IEEE/CVF Conference on Computer Vision and Pattern Recognition (CVPR)}, 2019, pp. 2647--2655.

\bibitem{otsu1979threshold}
N.~Otsu, ``A threshold selection method from gray-level histograms,'' \emph{IEEE transactions on systems, man, and cybernetics}, vol.~9, no.~1, pp. 62--66, 1979.

\bibitem{brigger2000b}
P.~Brigger, J.~Hoeg, and M.~Unser, ``B-spline snakes: a flexible tool for parametric contour detection,'' \emph{IEEE Transactions on image processing}, vol.~9, no.~9, pp. 1484--1496, 2000.

\bibitem{fukui1995edge}
K.~Fukui, ``Edge extraction method based on separability of image features,'' \emph{IEICE transactions on information and systems}, vol.~78, no.~12, pp. 1533--1538, 1995.

\bibitem{liu2012ultrasound}
S.~Liu, J.~Wei, B.~Feng, W.~Lu, B.~Denby, Q.~Fang, and J.~Dang, ``An anisotropic diffusion filter for reducing speckle noise of ultrasound images based on separability,'' in \emph{Proceedings of The 2012 Asia Pacific Signal and Information Processing Association Annual Summit and Conference}, 2012, pp. 1--4.

\bibitem{liu13h_interspeech}
S.~Liu, J.~Wei, X.~Wang, W.~Lu, Q.~Fang, and J.~Dang, ``An anisotropic diffusion filter based on multidirectional separability,'' in \emph{Interspeech 2013}, 2013, pp. 3187--3190.

\bibitem{niigaki2012circular}
H.~Niigaki, J.~Shimamura, and M.~Morimoto, ``Circular object detection based on separability and uniformity of feature distributions using bhattacharyya coefficient,'' in \emph{Proceedings of the 21st International Conference on Pattern Recognition (ICPR2012)}, 2012, pp. 2009--2012.

\bibitem{ohkawa2011fast}
Y.~Ohkawa, C.~H. Suryanto, and K.~Fukui, ``Fast combined separability filter for detecting circular objects.'' in \emph{MVA}, 2011, pp. 99--107.

\bibitem{CHEN2013iriseyelid}
\BIBentryALTinterwordspacing
Q.~Chen, K.~Mastumoto, and H.~Wu, ``Iris-eyelid separability filter for irises tracking,'' \emph{Procedia Computer Science}, vol.~22, pp. 1029--1037, 2013, 17th International Conference in Knowledge Based and Intelligent Information and Engineering Systems - KES2013. [Online]. Available: \url{https://www.sciencedirect.com/science/article/pii/S1877050913009812}
\BIBentrySTDinterwordspacing

\bibitem{chen2013irisoutline}
Q.~Chen, K.~Matsumoto, and H.~Wu, ``An iris outline tracker: For various eye shapes in a single video camera without infrared illumination,'' in \emph{2013 2nd IAPR Asian Conference on Pattern Recognition}, 2013, pp. 862--866.

\bibitem{kikuchi2014road}
I.~Kikuchi and K.~Onoguchi, ``Occluded side road detection using a single camera,'' in \emph{17th International IEEE Conference on Intelligent Transportation Systems (ITSC)}, 2014, pp. 2247--2248.

\bibitem{morita2016pupil}
Y.~Morita, H.~Takano, and K.~Nakamura, ``Pupil diameter measurement in visible-light environment using separability filter,'' in \emph{2016 IEEE International Conference on Systems, Man, and Cybernetics (SMC)}, 2016, pp. 000\,934--000\,939.

\bibitem{piegl1996nurbs}
L.~Piegl and W.~Tiller, \emph{The NURBS book}.\hskip 1em plus 0.5em minus 0.4em\relax Springer Science \& Business Media, 1996.

\bibitem{extremeclick}
D.~P. Papadopoulos, J.~R.~R. Uijlings, F.~Keller, and V.~Ferrari, ``Extreme clicking for efficient object annotation,'' in \emph{Proceedings of IEEE international conference on computer vision}, 2017, pp. 4940--4949.

\bibitem{deepextremecut}
K.-K. Maninis, S.~Caelles, J.~Pont-Tuset, and L.~{Van Gool}, ``Deep extreme cut: From extreme points to object segmentation,'' in \emph{IEEE Conference on Computer Vision and Pattern Recognition}, 2018.

\bibitem{octagonmask}
X.~Zhou, J.~Zhuo, and P.~Kr{\"a}henb{\"u}hl, ``Bottom-up object detection by grouping extreme and center points,'' in \emph{IEEE Conference on Computer Vision and Pattern Recognition}.

\end{thebibliography}
